\documentclass[11pt,a4paper]{article}
\usepackage[hyperref]{emnlp2020}
\usepackage{times}
\usepackage{latexsym}

\usepackage{microtype}

\usepackage{bm}
\usepackage{amsmath}
\usepackage{graphicx}
\usepackage{booktabs}
\usepackage{caption,subcaption}
\usepackage{multirow}
\usepackage{makecell}
\usepackage{enumitem}
\usepackage{amsfonts}

\aclfinalcopy 

\newcommand{\expnumber}[2]{{#1}\mathrm{e}{#2}}

\title{Learning to Learn to Disambiguate: \\ Meta-Learning for Few-Shot Word Sense Disambiguation}

\author{Nithin Holla \\
  ILLC, University of Amsterdam \\
  \texttt{nithin.holla7@gmail.com} \\\And
  Pushkar Mishra \\
  Facebook AI\\
  \texttt{pushkarmishra@fb.com} \\\AND
  Helen Yannakoudakis \\
  Dept. of Informatics, King's College London \\
  \texttt{helen.yannakoudakis@kcl.ac.uk} \\ \And
  Ekaterina Shutova \\
  ILLC, University of Amsterdam \\
  \texttt{e.shutova@uva.nl} \\}

\date{}

\begin{document}

\maketitle

\begin{abstract}
The success of deep learning methods hinges on the availability of large training datasets annotated for the task of interest. In contrast to human intelligence, these methods lack versatility and struggle to learn and adapt quickly to new tasks, where labeled data is scarce. Meta-learning aims to solve this problem by training a model on a large number of few-shot tasks, with an objective to learn new tasks quickly from a small number of examples. In this paper, we propose a meta-learning framework for few-shot word sense disambiguation (WSD), where the goal is to learn to disambiguate unseen words from only a few labeled instances. Meta-learning approaches have so far been typically tested in an $N$-way, $K$-shot classification setting where each task has $N$ classes with $K$ examples per class. Owing to its nature, WSD deviates from this controlled setup and requires the models to handle a large number of highly unbalanced classes. We extend several popular meta-learning approaches to this scenario, and analyze their strengths and weaknesses in this new challenging setting.
\end{abstract}

\section{Introduction}
Natural language is inherently ambiguous, with many words having a range of possible meanings. Word sense disambiguation (WSD) is a core task in natural language understanding, where the goal is to associate words with their correct contextual meaning from a pre-defined sense inventory. WSD has been shown to improve downstream tasks such as machine translation \citep{chan-wsd-translation} and information retrieval \citep{zhong-wsd-ir}. However, it is considered an AI-complete problem \citep{navigli_wsd_survey} -- it requires an intricate understanding of language as well as real-world knowledge.

Approaches to WSD typically rely on (semi-) supervised learning \citep{zhong-IMS,melamud-context2vec,kageback-wsd,yuan-semi_wsd} or are knowledge-based \citep{lesk_wsd,agirre-random_walk,moro-babelfy}. While supervised methods generally outperform the knowledge-based ones \citep{raganato-framework}, they require data manually annotated with word senses, which are expensive to produce at a large scale. These methods also tend to learn a classification model for each word independently, and hence may perform poorly on words that have a limited amount of annotated data. Yet, alternatives that involve a single supervised model for all words \citep{raganato-wsd} still do not adequately solve the problem for rare words \citep{kumar-zero_shot_wsd}. 

Humans, on the other hand, have a remarkable ability to learn from just a handful of examples \citep{lake-human_few_shot}. This inspired researchers to investigate techniques that would enable machine learning models to do the same. One such approach is transfer learning \citep{caruana_transfer_learning}, which aims to improve the models' data efficiency by transferring features between tasks. However, it still fails to generalize to new tasks in the absence of a considerable amount of task-specific data for fine-tuning \citep{yogatama}. Meta-learning, known as \textit{learning to learn} \citep{schmidhuber_thesis,bengio_metalearning,thrun_metalearning}, is an alternative paradigm that draws on past experience in order to learn and adapt to new tasks quickly: the model is trained on a number of related tasks such that it can solve unseen tasks using only a small number of training examples. A typical meta-learning setup consists of two components: a \textit{learner} that adapts to each task from its small training data; and a \textit{meta-learner} that guides the learner by acquiring knowledge that is common across all tasks.

Meta-learning has recently emerged as a promising approach to few-shot learning. It has achieved success in computer vision \citep{triantafillou_metadataset,fontanini-metalgan,hendryx-img_segmentation,wang-tracking} and reinforcement learning \citep{wang-rl,duan-rl2,alet-curiosity}. It has also recently made its way into NLP, and has been applied to machine translation \citep{gu-etal-2018-meta}, relation \citep{obamuyide-vlachos-2019-rel-classification} and text \citep{yu-diverse} classification, and sentence-level semantic tasks \citep{dou-etal-meta,bansal_meta}.

In this paper, we present the first meta-learning approach to WSD. We propose models that learn to rapidly disambiguate new words from only a few labeled examples. Owing to its nature, WSD exhibits inter-word dependencies within sentences, has a large number of classes, and inevitable class imbalances; all of which present new challenges compared to the controlled setup in most current meta-learning approaches. To address these challenges we extend three popular meta-learning algorithms to this task: prototypical networks \citep{Snell}, model-agnostic meta-learning (MAML) \citep{Finn} and a hybrid thereof -- ProtoMAML \citep{triantafillou_metadataset}. We investigate meta-learning using three underlying model architectures, namely recurrent networks, multi-layer perceptrons (MLP) and transformers \citep{vaswani-transformers}, and experiment with varying number of sentences available for task-specific fine-tuning.
We evaluate the model's rapid adaptation ability by testing on a set of new, unseen words, thus demonstrating its ability to learn new word senses from a small number of examples.

Since there are no few-shot WSD benchmarks available, we create a few-shot version of a publicly available WSD dataset. We release our code as well as the scripts used to generate our few-shot data setup to facilitate further research.\footnote{\url{https://github.com/Nithin-Holla/MetaWSD}}

\section{Related Work}
\subsection{Meta-learning}
In contrast to ``traditional'' machine learning approaches, meta-learning involves a different paradigm known as \textit{episodic learning}. The training and test sets in meta-learning are referred to as \textit{meta-training set} ($\mathcal{D}_{meta \text{-} train}$) and \textit{meta-test set} ($\mathcal{D}_{meta \text{-} test}$) respectively. Both sets consist of \textit{episodes} rather than individual data points. Each episode constitutes a task $\mathcal{T}_i$, comprising a small number of training examples for adaptation -- the \textit{support set} $\mathcal{D}_{support}^{(i)}$, and a separate set of examples for evaluation -- the \textit{query set} $\mathcal{D}_{query}^{(i)}$. A typical setup for meta-learning is the balanced $N$-way, $K$-shot setting where each episode has $N$ classes with $K$ examples per class in its support set.

Meta-learning algorithms are broadly categorized into three types: \textit{metric-based} \citep{Koch2015SiameseNN,Vinyals,sung2017_relationalnet,Snell}, \textit{model-based} \citep{santoro_mann,munkhdalai_metanet}, and \textit{optimization-based} \citep{Ravi,Finn,Nichol}. Metric-based methods first embed the examples in each episode into a high-dimensional space typically using a neural network. Next, they obtain the probability distribution over labels for all the query examples based on a kernel function that measures the similarity with the support examples. Model-based approaches aim to achieve rapid learning directly through their architectures. They typically employ external memory so as to remember key examples encountered in the past. Optimization-based approaches explicitly include generalizability in their objective function and optimize for the same. In this paper, we experiment with metric-based and optimization-based approaches, as well as a hybrid thereof.

\subsection{Meta-learning in NLP} \label{sec:meta_nlp}
Meta-learning in NLP is still in its nascent stages. \citet{gu-etal-2018-meta} apply meta-learning to the problem of neural machine translation where they meta-train on translating high-resource languages to English and meta-test on translating low-resource languages to English. \citet{obamuyide-vlachos-2019-rel-classification} use meta-learning for relation classification and \citet{obamuyide-vlachos-2019-rel-extract} for relation extraction in a lifelong learning setting. \citet{chen-relational} consider relation learning and apply meta-learning to few-shot link prediction in knowledge graphs. \citet{dou-etal-meta} perform meta-training on certain high-resource tasks from the GLUE benchmark \citep{wang-glue} and meta-test on certain low-resource tasks from the same benchmark. \citet{bansal_meta} propose a softmax parameter generator component that can enable a varying number of classes in the meta-training tasks. They choose the tasks in GLUE along with SNLI \citep{bowman_snli} for meta-training, and use entity typing, relation classification, sentiment classification, text categorization, and scientific NLI as the test tasks. Meta-learning has also been explored for few-shot text classification \citep{yu-diverse,geng-induction,jiang_text,sun-hierarchical}. \citet{wu-etal-multilabel} employ meta-reinforcement learning techniques for multi-label classification, with experiments on entity typing and text classification. \citet{hu-etal-vocab} use meta-learning to learn representations of out-of-vocabulary words, framing it as a regression task. 

\subsection{Supervised WSD}
Early supervised learning approaches to WSD relied on hand-crafted features extracted from the context words \citep{lee-eval, navigli_wsd_survey,zhong-IMS}. Later work used word embeddings as features for classification \citep{taghipour-wsd,rothe-autoextend,iacobacci-embeddings_wsd}. With the rise of deep learning, LSTM \citep{hochreiter_lstm} models became popular \citep{melamud-context2vec,kageback-wsd,yuan-semi_wsd}. While most work trained individual models per word, \citet{raganato-wsd} designed a single LSTM model with a large number of output units to disambiguate all words. \citet{peters-elmo} performed WSD by nearest neighbour matching with contextualized ELMo \citep{peters-elmo} embeddings. \citet{hadiwinoto-improved_wsd} used pre-trained contextualized representations from BERT \citep{devlin-bert} as features. GlossBERT \citep{huang-glossbert} incorporated sense definitions from WordNet \citep{miller-wordnet} to form context-gloss pairs while fine-tuning BERT for WSD. By taking WordNet's graph structure into account, EWISER \citep{bevilacqua-breaking} achieves the current state-of-the-art F1 score of 80.1\% on the benchmark by \citet{raganato-framework}.

\section{Task and Dataset} \label{sec:task_dataset}
We treat WSD as a few-shot word-level classification problem, where a sense is assigned to a word given its sentential context. As different words may have a different number of senses and sentences may have multiple ambiguous words, the standard setting of $N$-way, $K$-shot classification does not hold in our case. Specifically, different episodes can have a different number of classes and a varying number of examples per class -- a setting which is more realistic \citep{triantafillou_metadataset}.

\paragraph{Dataset} We use the SemCor corpus \citep{miller-semcor} manually annotated with senses from the New Oxford American Dictionary by \citet{yuan-semi_wsd}\footnote{\url{https://tinyurl.com/wsdcrp}. Note that the corpus does not have a standard train/validation/test split.}. With $37,176$ annotated sentences, this is one of the largest sense-annotated English corpora. We group the sentences in the corpus according to which word is to be disambiguated, and then randomly divide the words into disjoint meta-train, meta-validation and meta-test sets with a $60$:$15$:$25$ split. A sentence may have multiple occurrences of the same word, in which case we make predictions for all of them. We consider four different settings with the support set size $|S|$ = $4$, $8$, $16$ and $32$ sentences. The number of distinct words in the meta-training / meta-test sets is $985 / 270$, $985 / 259$, $799 / 197$ and $580 / 129$ respectively. The detailed statistics of the resulting dataset are shown in Appendix \ref{sec:data_stat}.

\paragraph{Training episodes} In the meta-training set, both the support and query sets have the same number of sentences. Our initial experiments using one word per episode during meta-training yielded poor results due to an insufficient number of episodes. To overcome this problem and design a suitable meta-training setup, we instead create episodes with multiple annotated words in them. Specifically, each episode consists of $r$ sampled words $\{z_j\}_{j=1}^r$ and $min(\lfloor |S| / r \rfloor, \nu(z_j))$ senses for each of those words, where $\nu(z_j)$ is the number of senses for word $z_j$. Therefore, each task in the meta-training set is the disambiguation of $r$ words between up to $|S|$ senses. We set $r = 2$ for $|S| = 4$ and $r=4$ for the rest. Sentences containing these senses are then sampled for the support and query sets such that the classes are as balanced as possible. For example, for $|S| = 8$, we first choose 4 words and 2 senses for each, and then sample one sentence for each word-sense pair. The labels for the senses are shuffled across episodes, i.e., one sense can have a different label when sampled in another episode. This is key in meta-learning as it prevents memorization \citep{yin-memorization}. The advantage of our approach for constructing meta-training episodes is that it allows for generating a combinatorially large number of tasks. Herein, we use a total number of $10,000$ meta-training episodes. 
\paragraph{Evaluation episodes} For the meta-validation and meta-test sets, each episode corresponds to the task of disambiguating a single word. While splitting the sentences into support and query sets, we ensure that senses in the query set are present in the support set.  Furthermore, we only consider words with two or more senses in their query set. The distribution of episodes across different number of senses is shown in Appendix \ref{sec:data_stat}. Note that, unlike the meta-training tasks, our meta-test tasks represent a natural data distribution, therefore allowing us to test our models in a realistic setting.

\section{Methods}

Our models consist of three components: an encoder that takes the words in a sentence as input and produces a contextualized representation for each of them, a hidden linear layer that projects these representations to another space, and an output linear layer that produces the probability distribution over senses. The encoder and the hidden layer are shared across all tasks -- we denote this block as $f_{\bm{\theta}}$ with shared parameters $\bm{\theta}$. The output layer is randomly initialized for each task $\mathcal{T}_i$ (i.e. episode) -- we denote this as $g_{\bm{\phi}_i}$ with parameters $\bm{\phi}_i$. $\bm{\theta}$ is meta-learned whereas $\bm{\phi}_i$ is independently learned for each task. 
\begin{figure*}[ht]
\small
    \centering
    \begin{subfigure}[b]{0.325\textwidth}
         \centering
         \includegraphics[width=\textwidth]{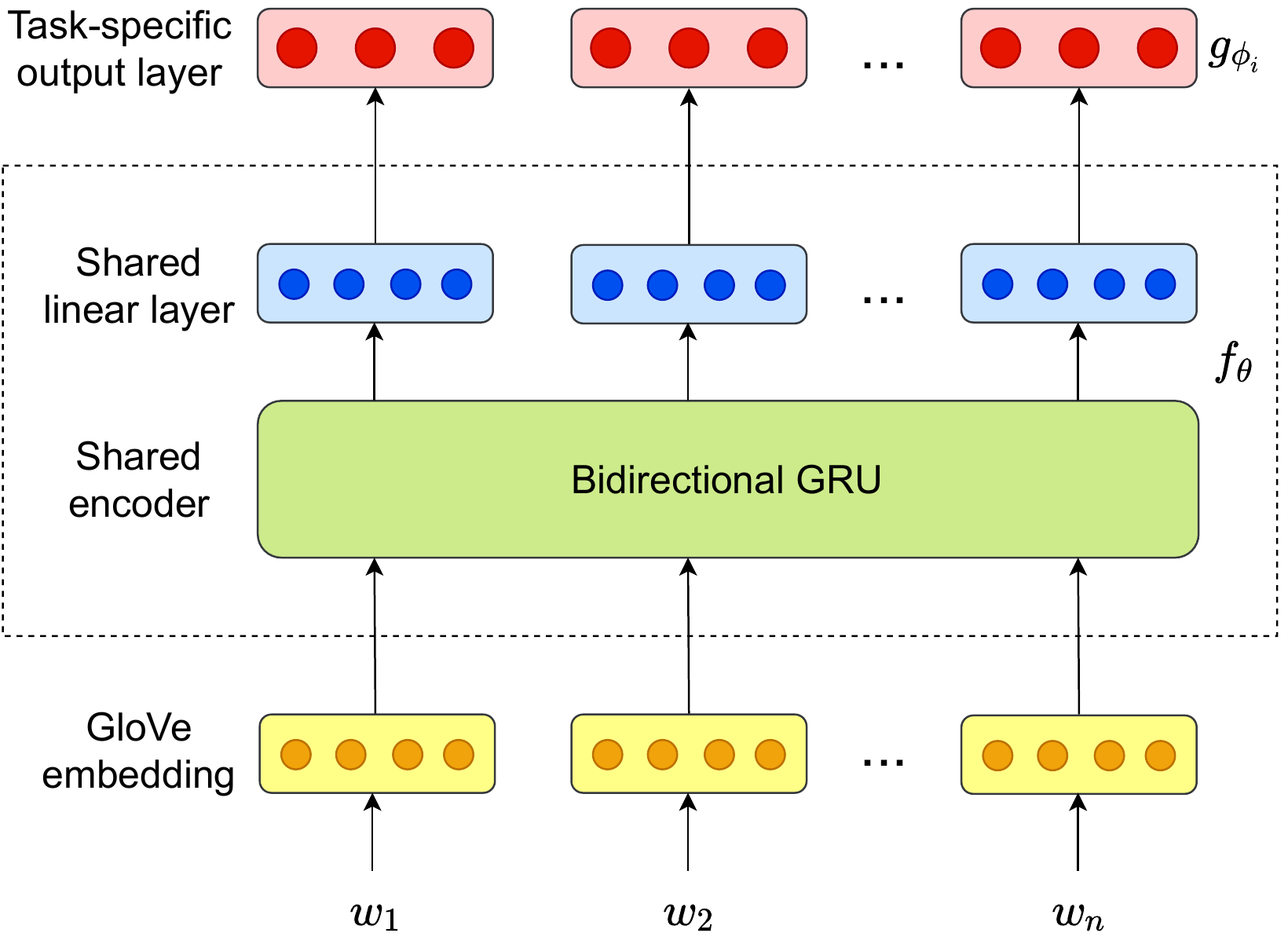}
         \caption{Bi-GRU encoder with GloVe input.}
         \label{fig:model_gru}
     \end{subfigure}
     \hfill
     \begin{subfigure}[b]{0.325\textwidth}
         \centering
         \includegraphics[width=\textwidth]{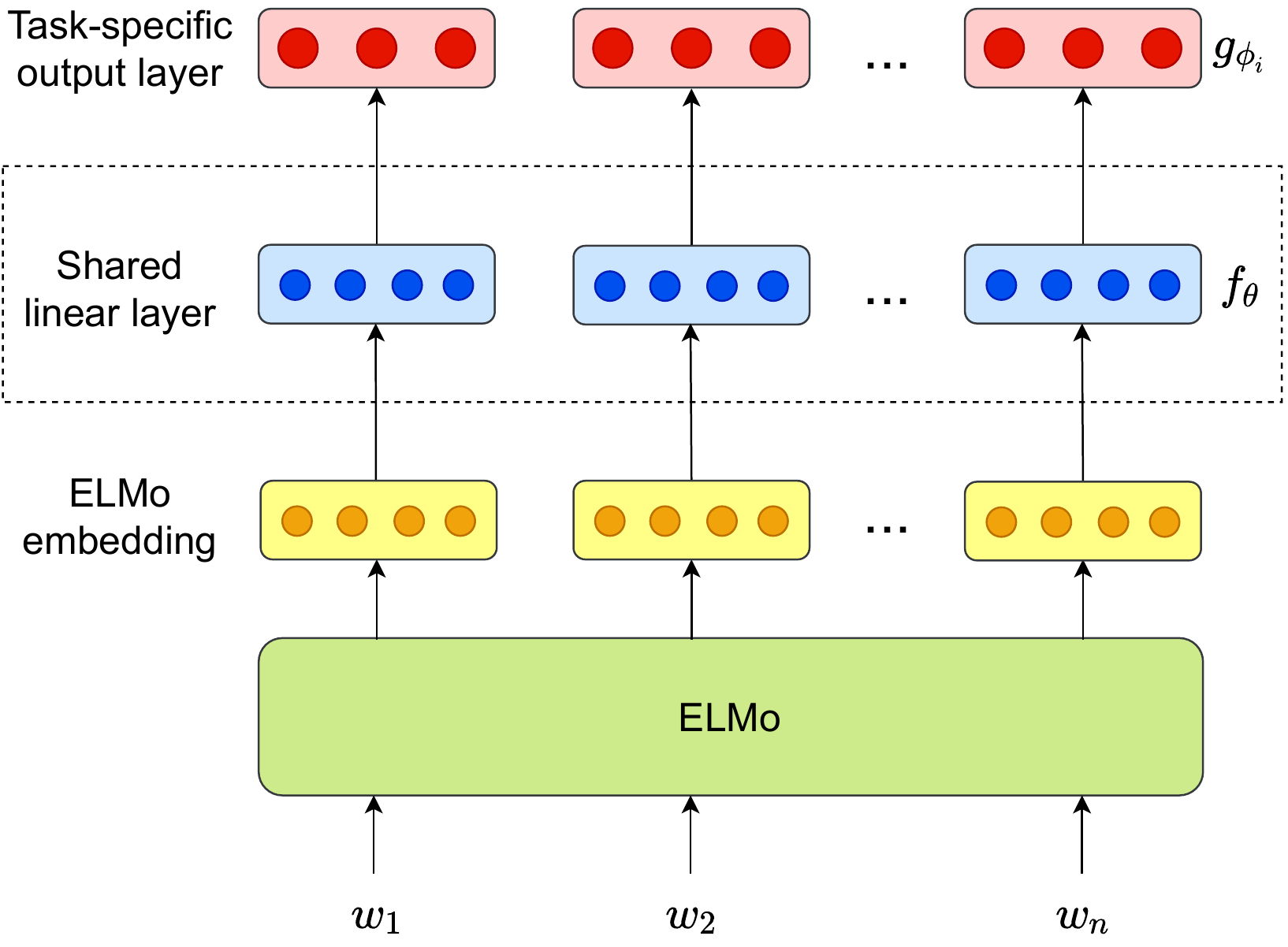}
         \caption{MLP with ELMo input.}
         \label{fig:model_elmo}
     \end{subfigure}
     \hfill
     \begin{subfigure}[b]{0.325\textwidth}
         \centering
         \includegraphics[width=\textwidth]{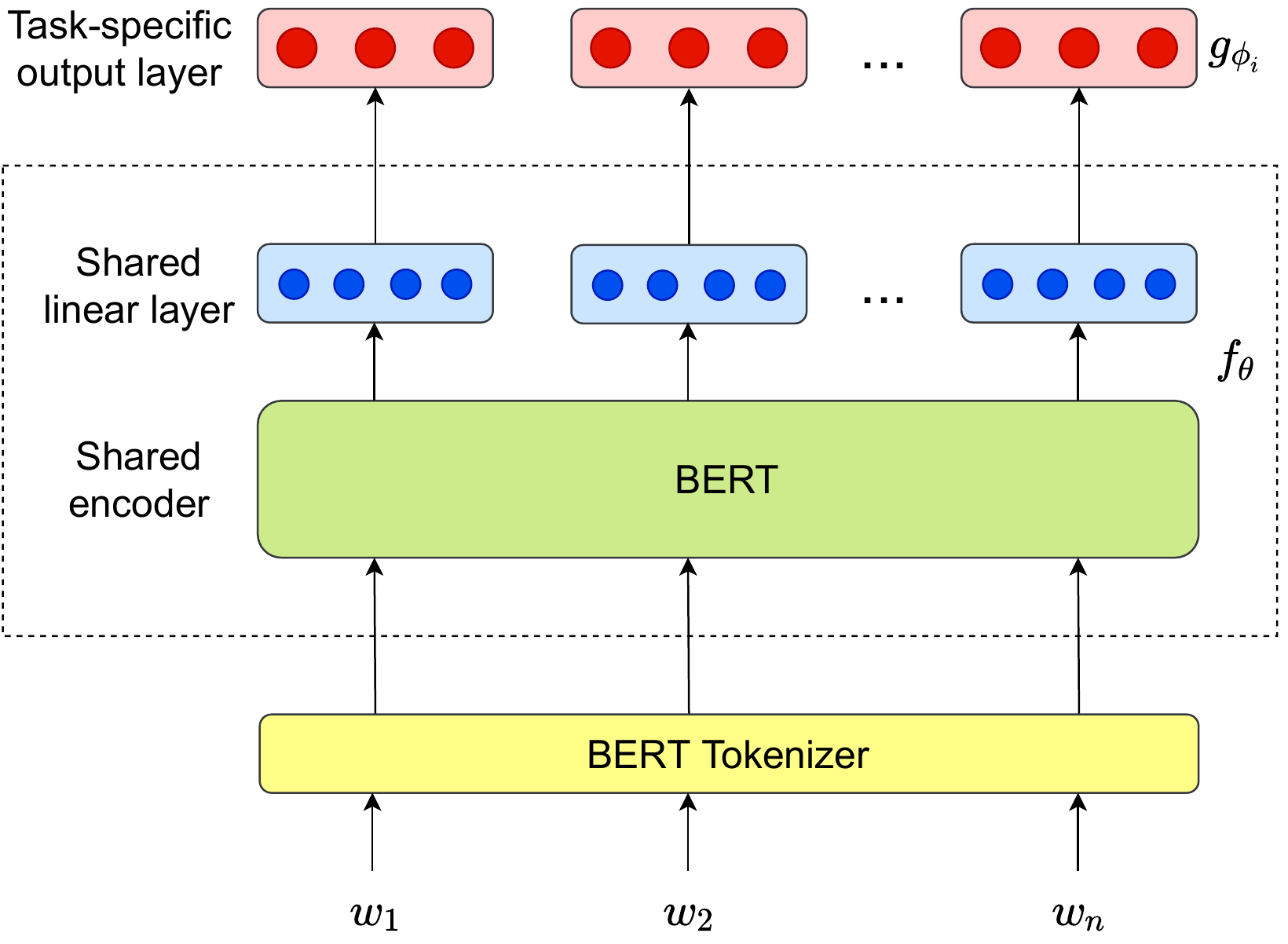}
         \caption{Entire BERT model as encoder.}
         \label{fig:model_bert}
     \end{subfigure}
    \caption{Model architecture showing the shared encoder, the shared linear layer and the task-specific linear layer. The inputs are words $w_1, w_2, ..., w_n$ of a sentence.}
    \label{fig:model}
\end{figure*}

\subsection{Model Architectures}
We experiment with three different architectures: (1) a single-layer bidirectional GRU \citep{cho-gru} with GloVe embeddings \citep{pennington-glove} as input that are not fine-tuned; (2) ELMo \citep{peters-elmo} embeddings that are not fine-tuned, followed by an MLP; and (3) $\text{BERT}_{\text{BASE}}$ \citep{devlin-bert} that is fine-tuned. The architecture of our three different models -- GloVe+GRU, ELMo+MLP and BERT -- is shown in Figure \ref{fig:model}. 

\subsection{Meta-learning Methods}

\paragraph{Prototypical Networks} Proposed by \citet{Snell}, prototypical networks is a metric-based approach. An embedding network $f_{\bm{\theta}}$ parameterized by $\bm{\theta}$ is used to produce a prototype vector for every class as the mean vector of the embeddings of all the support data points for that class. Suppose $S_c$ denotes the subset of the support set containing examples from class $c \in C$, the prototype $\bm{\mu_c}$ is:
\begin{align}
    \bm{\mu_c} &= \frac{1}{|S_c|} \sum_{\bm{x}_i \in S_c} f_{\bm{\theta}}(\bm{x}_i)
\end{align}
Given a distance function defined on the embedding space, the distribution over classes for a query point is calculated as a softmax over negative distances to the class prototypes.

We generate the prototypes (one per sense) from the output of the shared block $f_{\bm{\theta}}$ for the support examples. Instead of using $g_{\bm{\phi}_i}$, we obtain the probability distribution for the query examples based on the distance function. Parameters $\bm{\theta}$ are updated after every episode using the Adam optimizer \citep{kingma-adam}:
\begin{align}
    \bm{\theta} \leftarrow \text{Adam}(\mathcal{L}_{\mathcal{T}_i}^q, \bm{\theta}, \beta )
\end{align}
where $\mathcal{L}_{\mathcal{T}_i}^q$ is the cross-entropy loss on the query set and $\beta$ is the \textit{meta learning rate}.

\paragraph{Model-Agnostic Meta-Learning (MAML)} MAML \cite{Finn} is an optimization-based approach designed for the $N$-way, $K$-shot classification setting. The goal of optimization is to train a model's initial parameters such that it can perform well on a new task after only a few gradient steps on a small amount of data. Tasks are drawn from a distribution $p(\mathcal{T})$. The model's parameters are adapted from $\bm{\theta}$ to a task $\mathcal{T}_i$ using gradient descent on $D_{support}^{(i)}$ to yield $\bm{\theta}_i'$. This step is referred to as inner-loop optimization. With $m$ gradient steps, the update is: 
\begin{align} \label{eqn:inner_loop}
    \bm{\theta}_i' = U(\mathcal{L}_{\mathcal{T}_i}^s, \bm{\theta}, \alpha, m),
\end{align}
where $U$ is an optimizer such as SGD, $\alpha$ is the inner-loop learning rate
and $\mathcal{L}_{\mathcal{T}_i}^s$ is the loss for the task computed on $D_{support}^{(i)}$. The meta-objective is to have $f_{\bm{\theta}_i'}$ generalize well across tasks from $p(\mathcal{T})$:
\begin{align}
    J(\bm{\theta}) &= \sum_{\mathcal{T}_i \sim p(\mathcal{T})} \mathcal{L}_{\mathcal{T}_i}^q (f_{U(\mathcal{L}_{\mathcal{T}_i}^s, \bm{\theta}, \alpha, m)}).
\end{align}
where the loss $\mathcal{L}_{\mathcal{T}_i}^q$ is computed on $D_{query}^{(i)}$. The meta-optimization, or outer-loop optimization, does the update with the outer-loop learning rate $\beta$:
\begin{align}
    \bm{\theta} \leftarrow \bm{\theta} - \beta \nabla_{\bm{\theta}} \sum_{\mathcal{T}_i \sim p(\mathcal{T})} \mathcal{L}_{\mathcal{T}_i}^q (f_{\bm{\theta}_i'})
\end{align}
This involves computing second-order gradients, i.e., the backward pass works through the update step in Equation \ref{eqn:inner_loop} -- a computationally expensive process. \citet{Finn} propose a first-order approximation, called FOMAML, which computes the gradients with respect to $\bm{\theta}_i'$ rather than $\bm{\theta}$. The outer-loop optimization step thus reduces to:
\begin{align}
    \bm{\theta} \leftarrow \bm{\theta} - \beta \sum_{\mathcal{T}_i \sim p(\mathcal{T})} \nabla_{\bm{\theta}_i'} \mathcal{L}_{\mathcal{T}_i}^q (f_{\bm{\theta}_i'})
\end{align}

FOMAML does not generalize outside the $N$-way, $K$-shot setting, since it assumes a fixed number of classes across tasks. We therefore extend it with output layer parameters $\bm{\phi}_i$ that are adapted per task. During the inner-loop for each task, the optimization is performed as follows:
\begin{align} \label{eqn:new_inner_loop}
    \bm{\theta}_i', \bm{\phi}_i' \leftarrow \text{SGD}(\mathcal{L}_{\mathcal{T}_i}^s, \bm{\theta}, \bm{\phi}_i, \alpha, \gamma, m)
\end{align}
where $\alpha$ and $\gamma$ are the learning rates for the shared block and output layer respectively. We introduce different learning rates because the output layer is randomly initialized per task and thus needs to learn aggressively, whereas the shared block already has past information and can thus learn slower. We refer to $\alpha$ as the \textit{learner learning rate} and $\gamma$ as the \textit{output learning rate}. The outer-loop optimization uses Adam:
\begin{align} \label{eqn:new_outer_loop}
    \bm{\theta} \leftarrow \text{Adam} \left( \sum_i \mathcal{L}_{\mathcal{T}_i}^q(\bm{\theta}_i', \bm{\phi}_i'), \beta \right)
\end{align}
where the gradients of $\mathcal{L}_{\mathcal{T}_i}^q$ are computed with respect to $\bm{\theta}_i'$, $\beta$ is the \textit{meta learning rate}, and the sum over $i$ is for all tasks in the batch. 

\paragraph{ProtoMAML} \citet{Snell} show that with Euclidean distance metric, prototypical networks are equivalent to a linear model with the following parameters: $\bm{w_c} = 2 \bm{\mu_c}$ and $b_c = -\bm{\mu_c}^T \bm{\mu_c}$, where $\bm{w_c}$ and $b_c$ are the weights and biases for the output unit corresponding to class $c$. \citet{triantafillou_metadataset} combine the strengths of prototypical networks and MAML by initializing the final layer of the classifier in each episode with these prototypical network-equivalent weights and biases and continue to learn with MAML, thus proposing a hybrid approach referred to as ProtoMAML. Similarly, using FOMAML would yield ProtoFOMAML. While updating $\bm{\theta}$, they allow the gradients to flow through the linear layer initialization. 

We construct the prototypes from the output from $f_{\bm{\theta}}$ for the support examples. The parameters $\bm{\phi}_i$ are initialized as described above. The learning then proceeds as in (FO)MAML; the only difference being that $\gamma$ need not be too high owing to the good initialization. Proto(FO)MAML thus supports a varying number of classes per task. 

\subsection{Baseline Methods}

\paragraph{Majority-sense baseline} This baseline always predicts the most frequent sense in the support set. Hereafter, we refer to it as \textit{MajoritySenseBaseline}.

\paragraph{Nearest neighbor classifier} This model predicts the sense of a query instance as the sense of its nearest neighbor from the support set in terms of cosine distance. We perform nearest neighbor matching with the ELMo embeddings of the words as well as with their BERT outputs but not with GloVe embeddings since they are the same for all senses. We refer to this baseline as \textit{NearestNeighbor}.

\paragraph{Non-episodic training} It is a single model that is trained on all tasks without any distinction between them -- it merges support and query sets, and is trained using mini-batching. The output layer is thus not task-dependent and the number of output units is equal to the total number of senses in the dataset. The softmax at the output layer is taken only over the relevant classes within the mini-batch. Instead of $\bm{\phi}_i$ per task, we now have a single $\bm{\phi}$. During training, the parameters are updated per mini-batch as: 
\begin{align}
    \bm{\theta}, \bm{\phi} \leftarrow \text{Adam}(\mathcal{L}_{\mathcal{T}_i}, \bm{\theta}, \bm{\phi}, \alpha)
\end{align}
where $\alpha$ is the learning rate. During the meta-testing phase, we independently fine-tune the trained model on the support sets of each of the tasks (in an episodic fashion) as follows:
\begin{align}
    \bm{\theta}_i', \bm{\phi}_i' \leftarrow \text{SGD}(\mathcal{L}_{\mathcal{T}_i}, \bm{\theta}, \bm{\phi}, \alpha, \gamma, m)
\end{align}
where the loss is computed on the support examples, $\alpha$ is the \textit{learner learning rate} as before and $\gamma$ is the \textit{output learning rate}. We refer to this model as \textit{NE-Baseline}.

\paragraph{Episodic fine-tuning baseline} For each of the meta-learning methods, we include a variant that only performs meta-testing starting from a randomly initialized model. It is equivalent to training from scratch on the support examples of each episode. We prepend the prefix \textit{EF-} to denote this.

\section{Experiments and Results}
\subsection{Experimental setup}
We use the meta-validation set to choose the best hyperparameters for the models. The chosen evaluation metric is the average of the macro F1 scores across all words in the meta-validation set. We report the same metric on the meta-test set. We employ early stopping by terminating training if the metric does not improve over two epochs. The size of the hidden state in GloVe+GRU is $256$, and the size of the shared linear layer is $64$, $256$ and $192$ for GloVe+GRU, ELMo+MLP and BERT respectively. The shared linear layer's activation function is \textit{tanh} for GloVe+GRU, and \textit{ReLU} for ELMo+MLP and BERT. For FOMAML, ProtoFOMAML and ProtoMAML, the batch size is set to $16$ tasks. The output layer for these is initialized anew in every episode, whereas in NE-Baseline it has a fixed number of $5612$ units. We use the \texttt{higher} package \citep{grefenstette-meta} to implement the MAML variants.

\begin{table*}[ht]
\small
\centering
\begin{tabular}{llllll}
\toprule
\multirow{2}{*}{\makecell{\textbf{Embedding/} \\ \textbf{Encoder}}} & \multicolumn{1}{c}{\multirow{2}{*}{\textbf{Method}}} &  \multicolumn{4}{c}{\textbf{Average macro F1 score}} \\
                       & & \multicolumn{1}{c}{$|S| = 4$} & \multicolumn{1}{c}{$|S| = 8$}  & \multicolumn{1}{c}{$|S| = 16$} & \multicolumn{1}{c}{$|S| = 32$} \\ \midrule
- & MajoritySenseBaseline & 0.247 & 0.259 & 0.264 & 0.261 \\ \midrule
\multirow{7}{*}{GloVe+GRU} & NearestNeighbor & -- & -- & -- & -- \\
& NE-Baseline & 0.532 $\pm$ 0.007 & 0.507 $\pm$ 0.005 & 0.479 $\pm$ 0.004 & 0.451 $\pm$ 0.009 \\
& EF-ProtoNet & 0.522 $\pm$ 0.008 & 0.539 $\pm$ 0.009 & 0.538 $\pm$ 0.003 & 0.562 $\pm$ 0.005 \\
& EF-FOMAML & 0.376 $\pm$ 0.011 & 0.341 $\pm$ 0.002 & 0.321 $\pm$ 0.004 & 0.303 $\pm$ 0.005 \\
& EF-ProtoFOMAML & 0.519 $\pm$ 0.006 & 0.529 $\pm$ 0.010 & 0.540 $\pm$ 0.004 & 0.553 $\pm$ 0.009 \\
& ProtoNet & \textbf{0.579 $\pm$ 0.004} & 0.601 $\pm$ 0.003 & \textbf{0.633 $\pm$ 0.008} & \textbf{0.654 $\pm$ 0.004} \\
& FOMAML & 0.536 $\pm$ 0.007 & 0.418 $\pm$ 0.005 & 0.392 $\pm$ 0.007 & 0.375 $\pm$ 0.005 \\
& ProtoFOMAML & 0.577 $\pm$ 0.011 & \textbf{0.616 $\pm$ 0.005} & 0.626 $\pm$ 0.005 & 0.631 $\pm$ 0.008 \\ \midrule
\multirow{7}{*}{ELMo+MLP} & NearestNeighbor & 0.624 & 0.641 & 0.645 & 0.654 \\
& NE-Baseline & 0.624 $\pm$ 0.013 & 0.640 $\pm$ 0.012 & 0.633 $\pm$ 0.001 & 0.614 $\pm$ 0.008 \\
& EF-ProtoNet & 0.609 $\pm$ 0.008 & 0.635 $\pm$ 0.004 & 0.661 $\pm$ 0.004 &  0.683 $\pm$ 0.003 \\
& EF-FOMAML & 0.463 $\pm$ 0.004 & 0.414 $\pm$ 0.006 & 0.383 $\pm$ 0.003 & 0.352 $\pm$ 0.003 \\
& EF-ProtoFOMAML & 0.604 $\pm$ 0.004 & 0.621 $\pm$ 0.004 & 0.623 $\pm$ 0.008 & 0.611 $\pm$ 0.005 \\
& ProtoNet & 0.656 $\pm$ 0.006 & 0.688 $\pm$ 0.004 & 0.709 $\pm$ 0.006 & 0.731 $\pm$ 0.006 \\
& FOMAML & 0.642 $\pm$ 0.009 & 0.589 $\pm$ 0.010 & 0.587 $\pm$ 0.012 & 0.575 $\pm$ 0.016 \\
& ProtoFOMAML & \textbf{0.670 $\pm$ 0.005} & \textbf{0.700 $\pm$ 0.004} & \textbf{0.724 $\pm$ 0.003} & \textbf{0.737 $\pm$ 0.007} \\ \midrule
\multirow{7}{*}{BERT} & NearestNeighbor & 0.681 & 0.704 & 0.716 & 0.741 \\
& NE-Baseline & 0.467 $\pm$ 0.157 & 0.599 $\pm$ 0.023 & 0.539 $\pm$ 0.025 & 0.473 $\pm$ 0.015 \\
& EF-ProtoNet & 0.594 $\pm$ 0.008 & 0.655 $\pm$ 0.004 & 0.682 $\pm$ 0.005 & 0.721 $\pm$ 0.009 \\
& EF-FOMAML & 0.445 $\pm$ 0.009 & 0.522 $\pm$ 0.007 & 0.450 $\pm$ 0.008 & 0.393 $\pm$ 0.002 \\
& EF-ProtoFOMAML & 0.618 $\pm$ 0.013 & 0.662 $\pm$ 0.006 & 0.654 $\pm$ 0.009 & 0.665 $\pm$ 0.009 \\
& ProtoNet & 0.696 $\pm$ 0.011 & 0.750 $\pm$ 0.008 & \textbf{0.755 $\pm$ 0.002} & \textbf{0.766 $\pm$ 0.003} \\
& FOMAML & 0.676 $\pm$ 0.018 & 0.550 $\pm$ 0.011 & 0.476 $\pm$ 0.010 & 0.436 $\pm$ 0.014 \\
& ProtoFOMAML & \textbf{0.719 $\pm$ 0.005} & \textbf{0.756 $\pm$ 0.007} & 0.744 $\pm$ 0.007 & 0.761 $\pm$ 0.005 \\
\bottomrule                       
\end{tabular}
\caption{Average macro F1 scores of the meta-test words.}
\label{tab:results}
\end{table*}

\begin{table*}[ht]
\small
\centering
\begin{tabular}{llllll}
\toprule
\multirow{2}{*}{\makecell{\textbf{Embedding/} \\ \textbf{Encoder}}} & \multicolumn{1}{c}{\multirow{2}{*}{\textbf{Method}}} & \multicolumn{4}{c}{\textbf{Average macro F1 score}} \\
                       & & \multicolumn{1}{c}{$|S| = 4$} & \multicolumn{1}{c}{$|S| = 8$}  & \multicolumn{1}{c}{$|S| = 16$} & \multicolumn{1}{c}{$|S| = 32$} \\ \midrule
\multirow{3}{*}{GloVe+GRU} & ProtoNet & \textbf{0.579 $\pm$ 0.004} & 0.601 $\pm$ 0.003 & \textbf{0.633 $\pm$ 0.008} & \textbf{0.654 $\pm$ 0.004} \\
& ProtoFOMAML & 0.577 $\pm$ 0.011 & 0.616 $\pm$ 0.005 & 0.626 $\pm$ 0.005 & 0.631 $\pm$ 0.008 \\
& ProtoMAML & \textbf{0.579 $\pm$ 0.006} & \textbf{0.617 $\pm$ 0.005} & 0.629 $\pm$ 0.006 & 0.633 $\pm$ 0.006 \\ \midrule
\multirow{3}{*}{ELMo+MLP} & ProtoNet & 0.656 $\pm$ 0.006 & 0.688 $\pm$ 0.004 & 0.709 $\pm$ 0.006 & 0.731 $\pm$ 0.006 \\
& ProtoFOMAML & 0.670 $\pm$ 0.005 & 0.700 $\pm$ 0.004 & \textbf{0.724 $\pm$ 0.003} & \textbf{0.737 $\pm$ 0.007} \\
& ProtoMAML & \textbf{0.671 $\pm$ 0.006} & \textbf{0.702 $\pm$ 0.006} & 0.722 $\pm$ 0.004 & 0.735 $\pm$ 0.008 \\ \midrule
\multirow{3}{*}{BERT} & ProtoNet & 0.696 $\pm$ 0.011 & \textbf{0.750 $\pm$ 0.008} & \textbf{0.755 $\pm$ 0.002} & \textbf{0.766 $\pm$ 0.003} \\
& ProtoFOMAML* & \textbf{0.697 $\pm$ 0.013} & \textbf{0.750 $\pm$ 0.005} & 0.741 $\pm$ 0.007 & 0.754 $\pm$ 0.006 \\
& ProtoMAML* & 0.690 $\pm$ 0.003 & 0.736 $\pm$ 0.004 & 0.737 $\pm$ 0.006 & 0.752 $\pm$ 0.006 \\
\bottomrule                       
\end{tabular}
\caption{Average macro F1 scores of the meta-test words for second-order gradient model variants as well as ProtoNet. (*Only the top layer fine-tuned and for only one inner-loop step)}
\label{tab:results2}
\end{table*}

\subsection{Results} 

In Table \ref{tab:results}, we report macro F1 scores averaged over all words in the meta-test set. We report the mean and standard deviation from five independent runs. We note that the results are not directly comparable across $|S|$ setups as, by their formulation, they involve different meta-test episodes.  

\paragraph{GloVe+GRU} All meta-learning methods perform better than their EF counterparts, indicating successful learning from the meta-training set. FOMAML fails to outperform NE-Baseline as well as the EF versions of the other meta-learning methods when $|S| = 8, 16, 32$. Interestingly, solely meta-testing is often better than NE-Baseline model which shows that the latter does not effectively transfer knowledge from the meta-training set. ProtoNet is the best-performing model (except when $|S|=8$), with ProtoMAML being a close second.  

\paragraph{ELMo+MLP} The scores for NearestNeighbor, NE-Baseline and the EF methods are higher compared to GloVe-based models, which can be attributed to the input embeddings being contextual. ProtoNet and ProtoFOMAML still improve over their EF counterparts due to meta-training. ProtoFOMAML outperforms other methods for all $|S|$, and FOMAML is comparatively weak.  

\paragraph{BERT} The scores for all methods are higher than in case of the previous architectures, except for NE-Baseline and FOMAML. BERT-based ProtoNet, as well as ProtoFOMAML, outperform all other approaches for all $|S|$. Furthermore, ProtoFOMAML is superior to ProtoNet for $|S| = 4, 8$ and vice versa for $|S| = 16, 32$. Overall, across architectures, we see that NE-Baseline and FOMAML consistently underperform, whereas ProtoNet and ProtoFOMAML are the most effective methods. Moreover, they achieve a high disambiguation performance with as few as 4 training examples, which in many cases approaches a one-shot classification setting for individual senses (see Appendix \ref{sec:data_stat}). The models are also relatively stable as indicated by the low standard deviations across runs.

\paragraph{Effect of second-order gradients} We further experiment with ProtoMAML, including second-order gradients. In Table \ref{tab:results2}, we report its F1 scores alongside ProtoNet and ProtoFOMAML. For BERT, we train ProtoMAML while fine-tuning only the top layer and only for one inner-loop update step due to its high computational cost. We also train an equivalent ProtoFOMAML variant for a fair comparison. We can observe that ProtoMAML obtains scores similar to ProtoFOMAML in most cases, indicating the effectiveness of the first-order approximation. ProtoFOMAML achieves higher scores than ProtoMAML in some cases, perhaps due to an overfitting effect induced by the latter. In light of these results, we argue that first-order ProtoFOMAML suffices for this task.

\subsection{Analysis}

\begin{figure}[t!]
    \centering
    \includegraphics[width=0.48\textwidth]{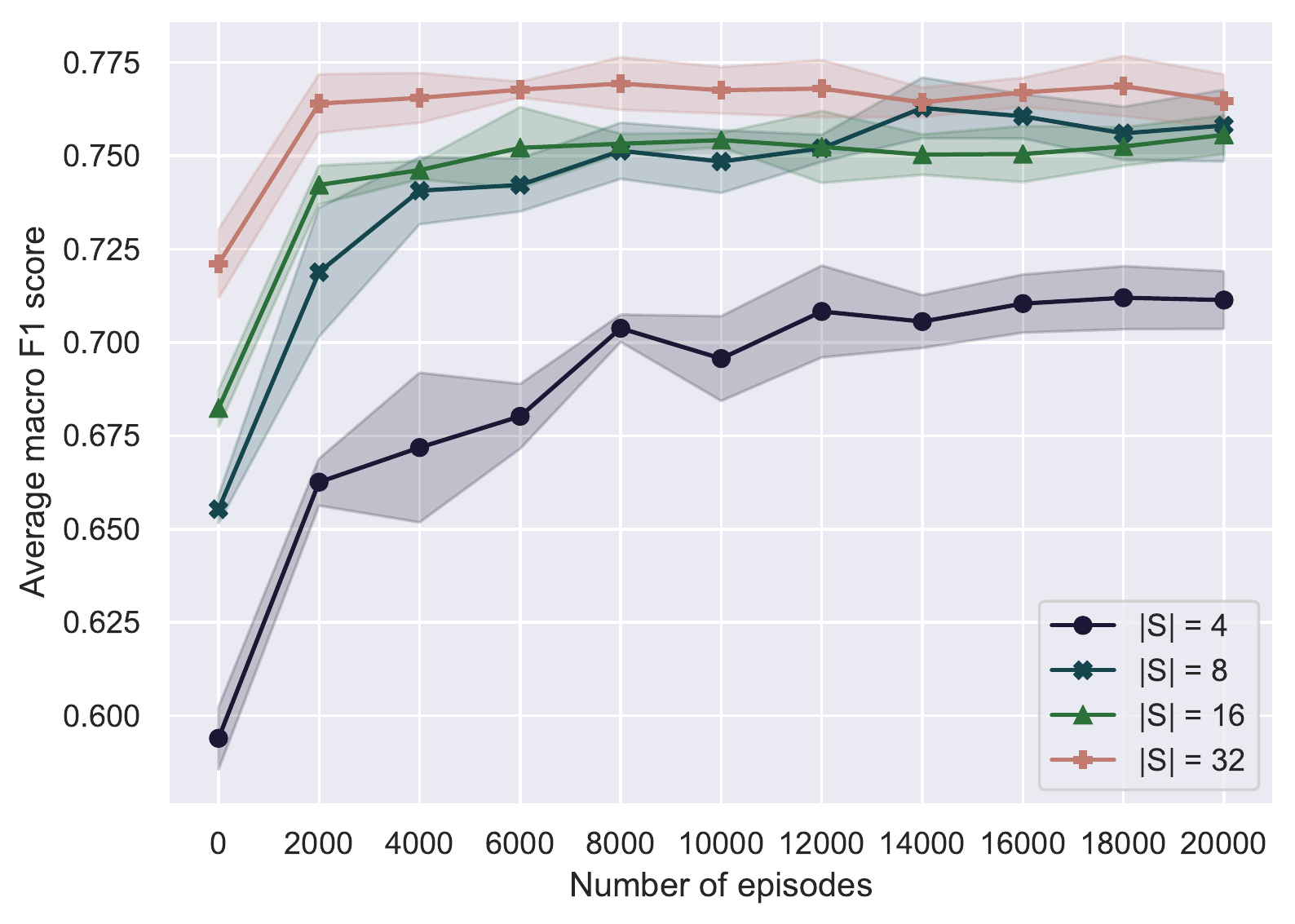}
    \caption{Average macro F1 score of ProtoNet+BERT as the number of meta-training episodes increases.}
    \label{fig:episode_exp}
\end{figure}

\paragraph{Effect of number of episodes} We first investigate whether using more meta-training episodes always translates to higher performance. We plot the average macro F1 score for one of our high-scoring models -- ProtoNet with BERT -- as the number of meta-training episodes increases (Figure \ref{fig:episode_exp}). The shaded region shows one standard deviation from the mean, obtained over five runs. Different $|S|$ setups reach peaks at different data sizes; however, overall, the largest gains come with a minimum of around $8,000$ episodes. 

\begin{figure*}[ht]
    \centering
    \begin{subfigure}[b]{0.24\textwidth}
         \centering
         \includegraphics[width=\textwidth]{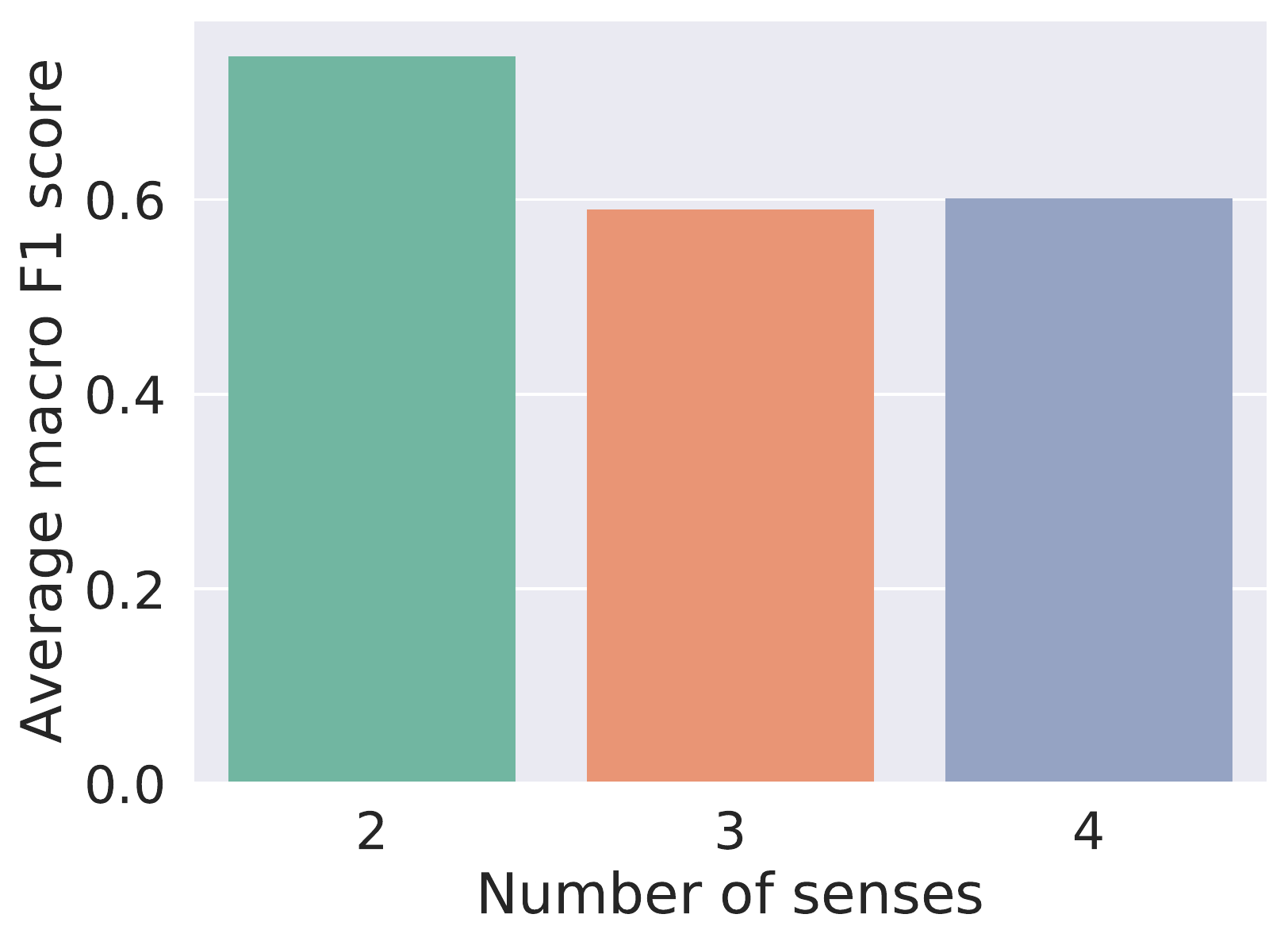}
         \caption{$|S| = 4$}
         \label{fig:sense_f1_4}
     \end{subfigure}
     \hfill
    \begin{subfigure}[b]{0.24\textwidth}
         \centering
         \includegraphics[width=\textwidth]{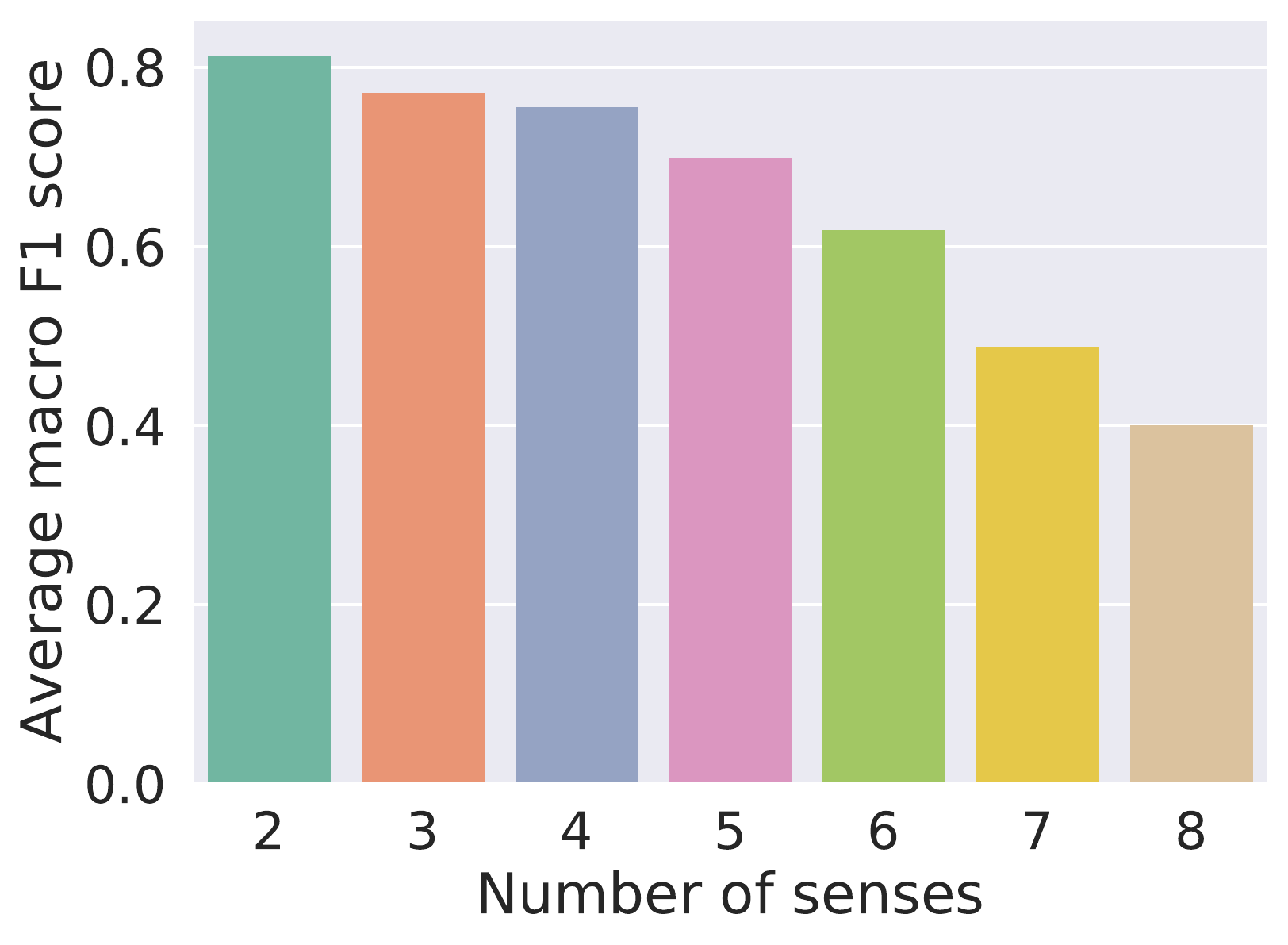}
         \caption{$|S| = 8$}
         \label{fig:sense_f1_8}
     \end{subfigure}
     \hfill
     \begin{subfigure}[b]{0.24\textwidth}
         \centering
         \includegraphics[width=\textwidth]{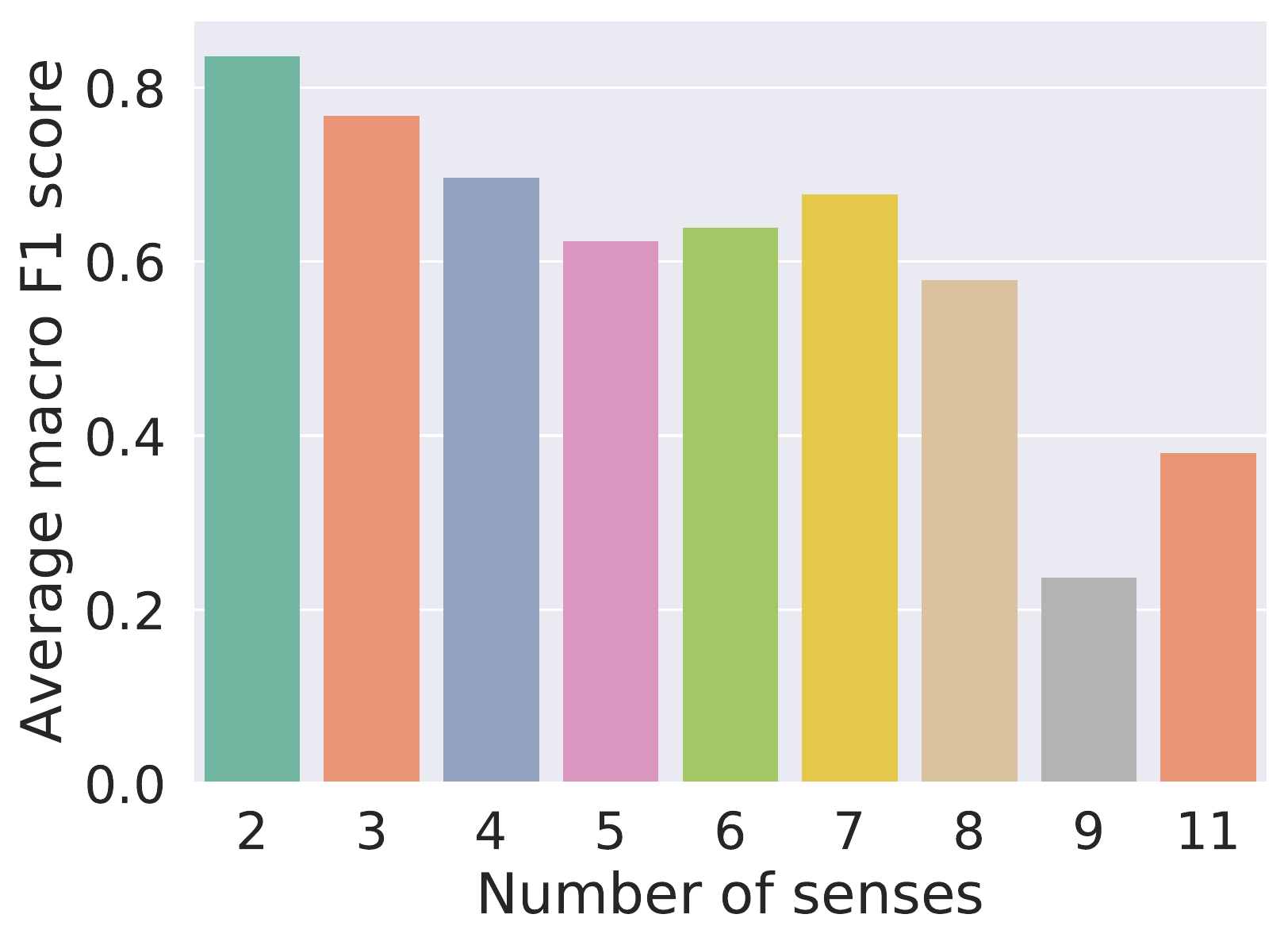}
         \caption{$|S| = 16$}
         \label{fig:sense_f1_16}
     \end{subfigure}
     \hfill
     \begin{subfigure}[b]{0.24\textwidth}
         \centering
         \includegraphics[width=\textwidth]{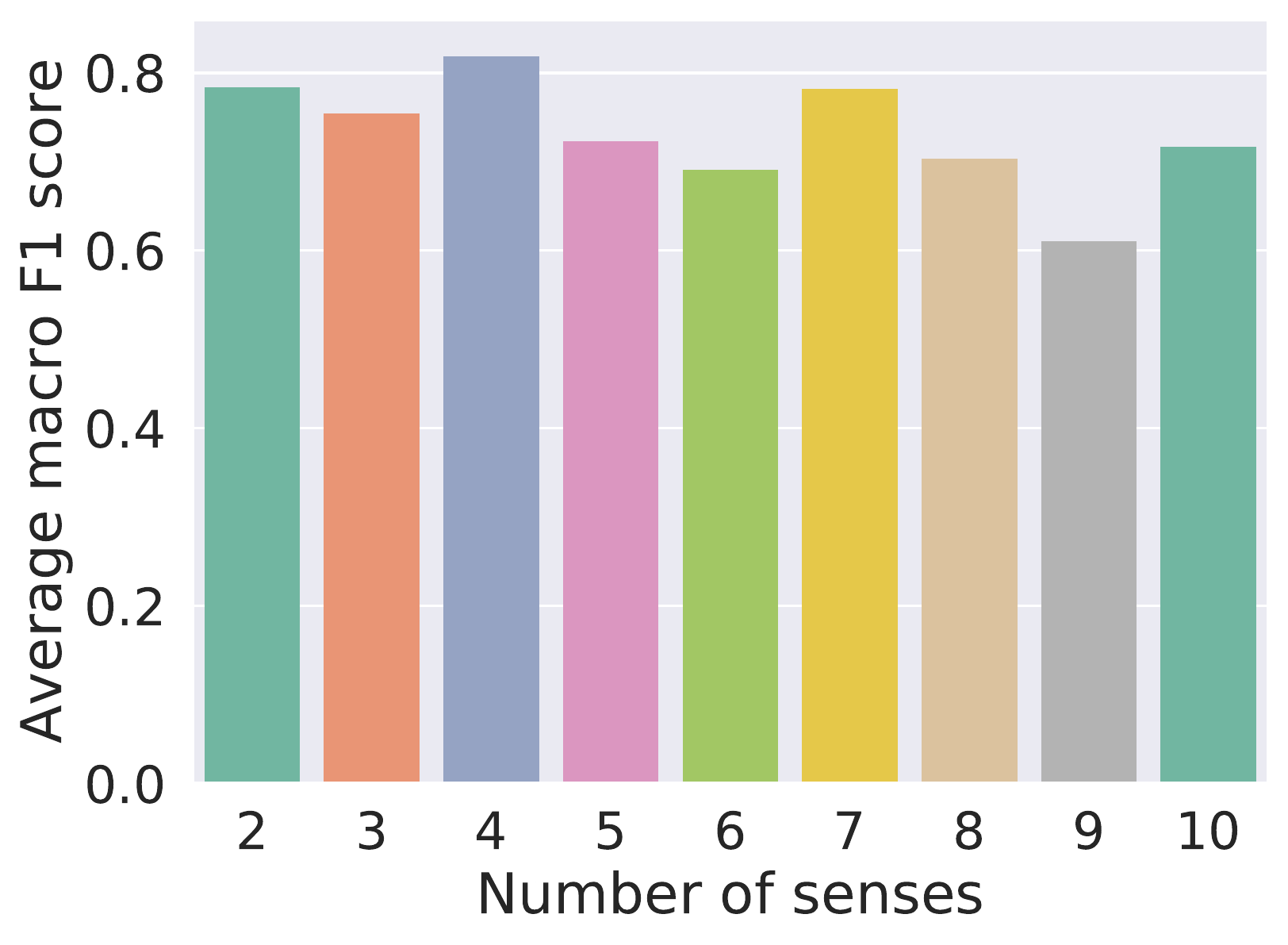}
         \caption{$|S| = 32$}
         \label{fig:sense_f1_32}
     \end{subfigure}
    \caption{Bar plot of macro F1 scores averaged over words with a given number of senses.}
    \label{fig:sense_f1}
\end{figure*}

\paragraph{Effect of number of senses} To investigate the variation in performance with the number of senses, in Figure \ref{fig:sense_f1}, we plot the macro F1 scores obtained from ProtoNet with BERT, averaged over words with a given number of senses in the meta-test set. We see a trend where the score reduces as the number of senses increase. Words with more senses seem to benefit from a higher support set size. For a word with $8$ senses, the $|S| = 32$ case is roughly a $4$-shot problem whereas it is roughly a $2$-shot and $1$-shot problem for $|S| = 16$ and $|S| = 8$ respectively. In this view, the disambiguation of words with many senses improves with $|S|$ due to an increase in the effective number of shots.

\paragraph{Challenging cases} Based on the 10 words that obtain the lowest macro F1 scores with ProtoNet with GloVe+GRU (Appendix \ref{sec:challenging_cases}), we see that verbs are the most challenging words to disambiguate without the advantage of pre-trained models and their disambiguation improves as $|S|$ increases.

\section{Discussion}

\begin{figure}[ht]
    \centering
     \begin{subfigure}[b]{0.8\columnwidth}
         \centering
         \includegraphics[width=\linewidth]{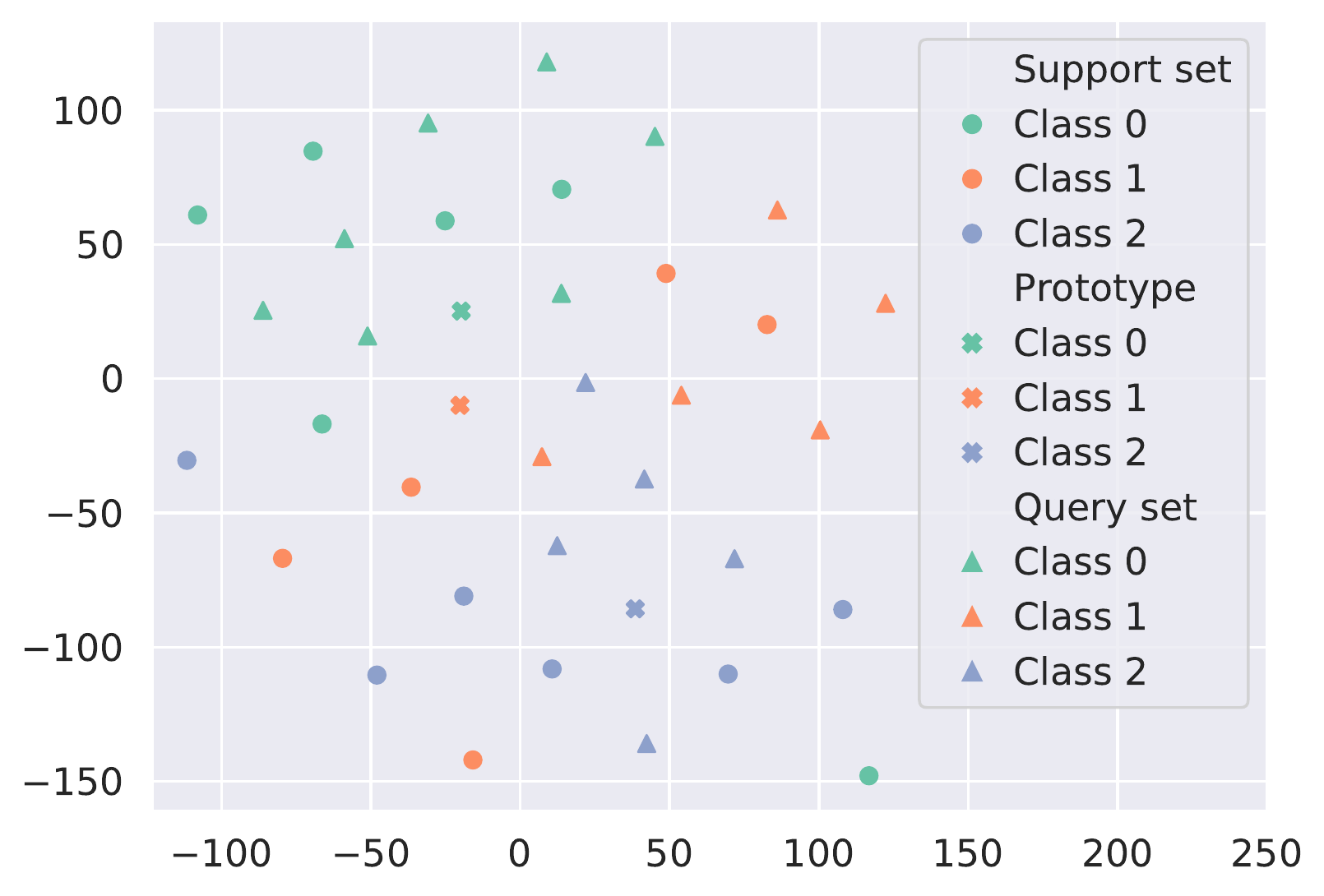}
         \caption{}
         \label{fig:elmo_visualize_field_sample}
     \end{subfigure} 
     \hfill
     \begin{subfigure}[b]{0.8\columnwidth}
         \centering
         \includegraphics[width=\linewidth]{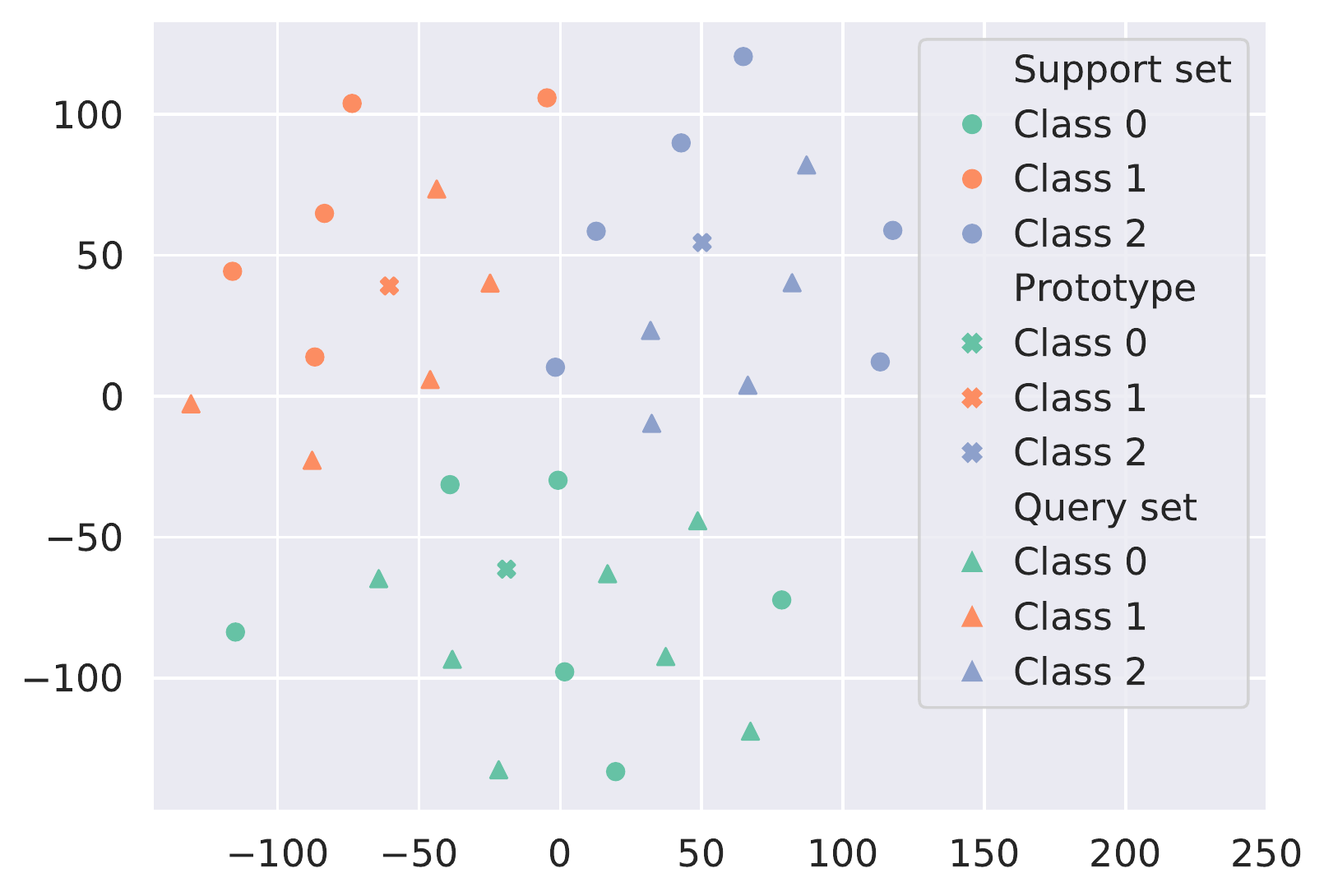}
         \caption{}
         \label{fig:proto_elmo_visualize_field_sample}
     \end{subfigure}
    \caption{t-SNE visualizations comparing ELMo embeddings (left) against representations generated by ProtoNet with ELMo+MLP (right) for the word `field'.}
    \label{fig:tsne_elmo_sample}
\end{figure}

Our results demonstrate that meta-learning outperforms the corresponding models trained in a non-episodic fashion when applied in a few-shot learning setting -- a finding consistent for all $|S|$ setups. Using the BERT-based models, we obtain up to $72\%$ average macro F1 score with as few as $4$ examples, and closely approach the reported state-of-the-art performance\footnote{Not a direct comparison due to different data splits.} with $|S|=\{16,32\}$.

The success of meta-learning is particularly evident with GloVe+GRU. GloVe embeddings are sense-agnostic and yet, ProtoNet, ProtoFOMAML and ProtoMAML approach the performance of some ELMo-based models, which enjoy the benefit of contextualization via large-scale pretraining.

Although contextualized representations from ELMo and BERT already contain information relevant to our task, integrating them into a meta-learning framework allows these models to substantially improve performance. To illustrate the advantage that meta-learning brings, we provide example t-SNE visualizations \citep{maaten_tsne} of the original ELMo embeddings and those generated by ProtoNet based on ELMo (Figure \ref{fig:tsne_elmo_sample}). The representations from ProtoNet are more accurately clustered with respect to the senses than the original ELMo representations. ProtoNet thus effectively learns to disambiguate new words, i.e. separate the senses into clusters, thereby improving upon ELMo embeddings. We provide further t-SNE visualizations in Appendix \ref{sec:tsne}.

The success of ProtoNet and ProtoFOMAML can be in part attributed to the nature of the problem -- WSD lends itself well to modeling approaches based on similarity \citep{navigli_wsd_survey,peters-elmo}. Their relative ranking, however, depends on the architecture and the value of $|S|$. ELMo+MLP has the simplest architecture and ProtoFOMAML -- an optimization-based method -- performs best. For GloVe+GRU and BERT, which are more complex architectures, lower-shot settings benefit from ProtoFOMAML and higher-shot settings from ProtoNet. The reasons for this effect, however, remain to be investigated in future work. 

Our experiments further highlight the weakness of FOMAML when applied beyond the $N$-way, $K$-shot setting. This may be due to the fact that the number of ``new'' output parameters in each episode is much greater than the number of support examples. Informed output layer initialization in Proto(FO)MAML is therefore important for effective learning in such scenarios. A similar problem with FOMAML is also pointed out by \citet{bansal_meta}, who design a differentiable parameter generator for the output layer.
 
\section{Conclusion}
Few-shot learning is a key capability for AI to reach human-like performance. The development of meta-learning methods is a promising step in this direction. We demonstrated the ability of meta-learning to disambiguate new words when only a handful of labeled examples are available. Given the data scarcity in WSD and the need for few-shot model adaptation to specific domains, we believe that meta-learning can yield a more general and widely applicable disambiguation model than traditional approaches. Interesting avenues to explore further would be a generalization of our models to disambiguation in different domains, to a multilingual scenario or to an altogether different task. 

\bibliography{main}
\bibliographystyle{acl_natbib}

\appendix

\section{Appendix}

\subsection{Data statistics} \label{sec:data_stat}

\begin{table*}[ht]
\small
    \centering
    \begin{tabular}{@{}cccccc@{}}
    \toprule
    \makecell{\textbf{Support} \\ \textbf{sentences}} & \textbf{Split} & \makecell{\textbf{No. of} \\ \textbf{words}} & \makecell{\textbf{No. of} \\ \textbf{episodes}} & \makecell{\textbf{No. of unique} \\ \textbf{sentences}} & \makecell{\textbf{Average no. of} \\ \textbf{senses}} \\ \midrule
    \multirow{3}{*}{4}  & Meta-training & 985 & 10000 & 27640 & 2.96 \\
                        & Meta-validation & 166 & 166 & 1293 & 2.60 \\
                        & Meta-test & 270 & 270 & 2062 & 2.60 \\ \midrule
    \multirow{3}{*}{8}  & Meta-training & 985 & 10000 & 27640 & 2.96 \\
                        & Meta-validation & 163 & 163 & 2343 & 3.06 \\
                        & Meta-test & 259 & 259 & 3605 & 3.16 \\ \midrule
    \multirow{3}{*}{16} & Meta-training & 799 & 10000 & 27973 & 3.07 \\
                        & Meta-validation & 146 & 146 & 3696 & 3.53 \\
                        & Meta-test & 197 & 197 & 4976 & 3.58 \\ \midrule
    \multirow{3}{*}{32} & Meta-training & 580 & 10000 & 27046 & 3.34 \\
                        & Meta-validation & 85 & 85 & 4129 & 3.94 \\
                        & Meta-test & 129 & 129 & 5855 & 3.52 \\
    \bottomrule
    \end{tabular}
    \caption{Statistics of our few-shot WSD dataset.}
    \label{tab:data_split}
\end{table*}

\begin{table}[ht]
\small
\centering
\begin{tabular}{p{0.08\textwidth}p{0.34\textwidth}}
\toprule
\multirow{2}{*}{$|S| = 4$} & ward, delicate, jam, lose, year, bounce, haul, introduce, guard, suffer \\ \midrule
\multirow{2}{*}{$|S| = 8$} & bad, work, give, clear, settle, bloom, draw, check, break, gather \\ \midrule
\multirow{2}{*}{$|S| = 16$} & move, appearance, in, green, fix, establishment, note, drag, cup, bounce \\ \midrule 
\multirow{2}{*}{$|S| = 32$} & independent, gather, north, square, do, bond, proper, pull, problem, language \\
\bottomrule
\end{tabular}
\caption{Words with the lowest macro F1 scores for ProtoNet with GloVe+GRU.}
\label{tab:worst_words}
\end{table}

\begin{figure*}[ht]
    \centering
    \begin{subfigure}[b]{0.24\textwidth}
         \centering
         \includegraphics[width=\textwidth]{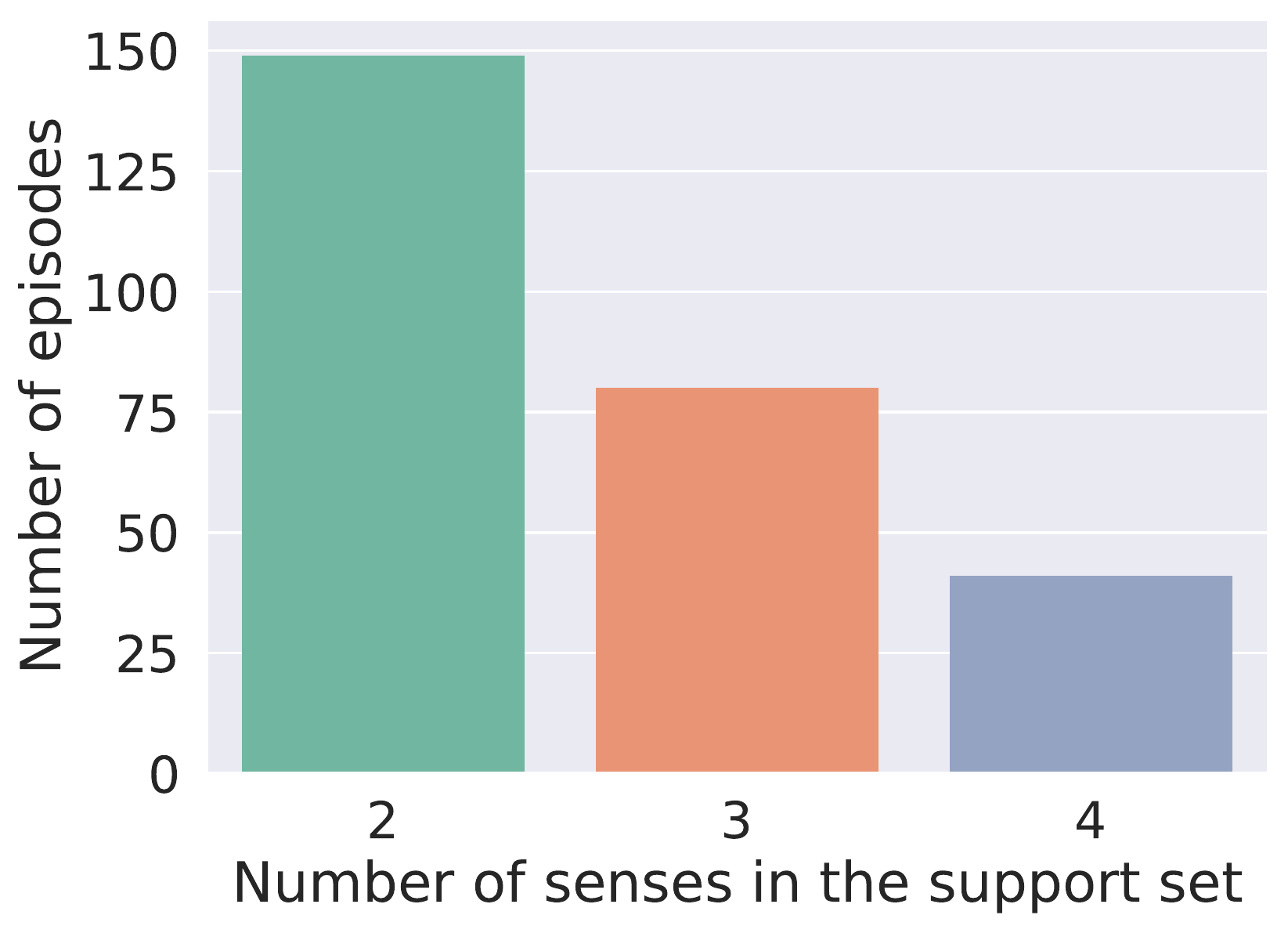}
         \caption{$|S| = 4$}
         \label{fig:sense_sepisode_4}
     \end{subfigure}
     \hfill
    \begin{subfigure}[b]{0.24\textwidth}
         \centering
         \includegraphics[width=\textwidth]{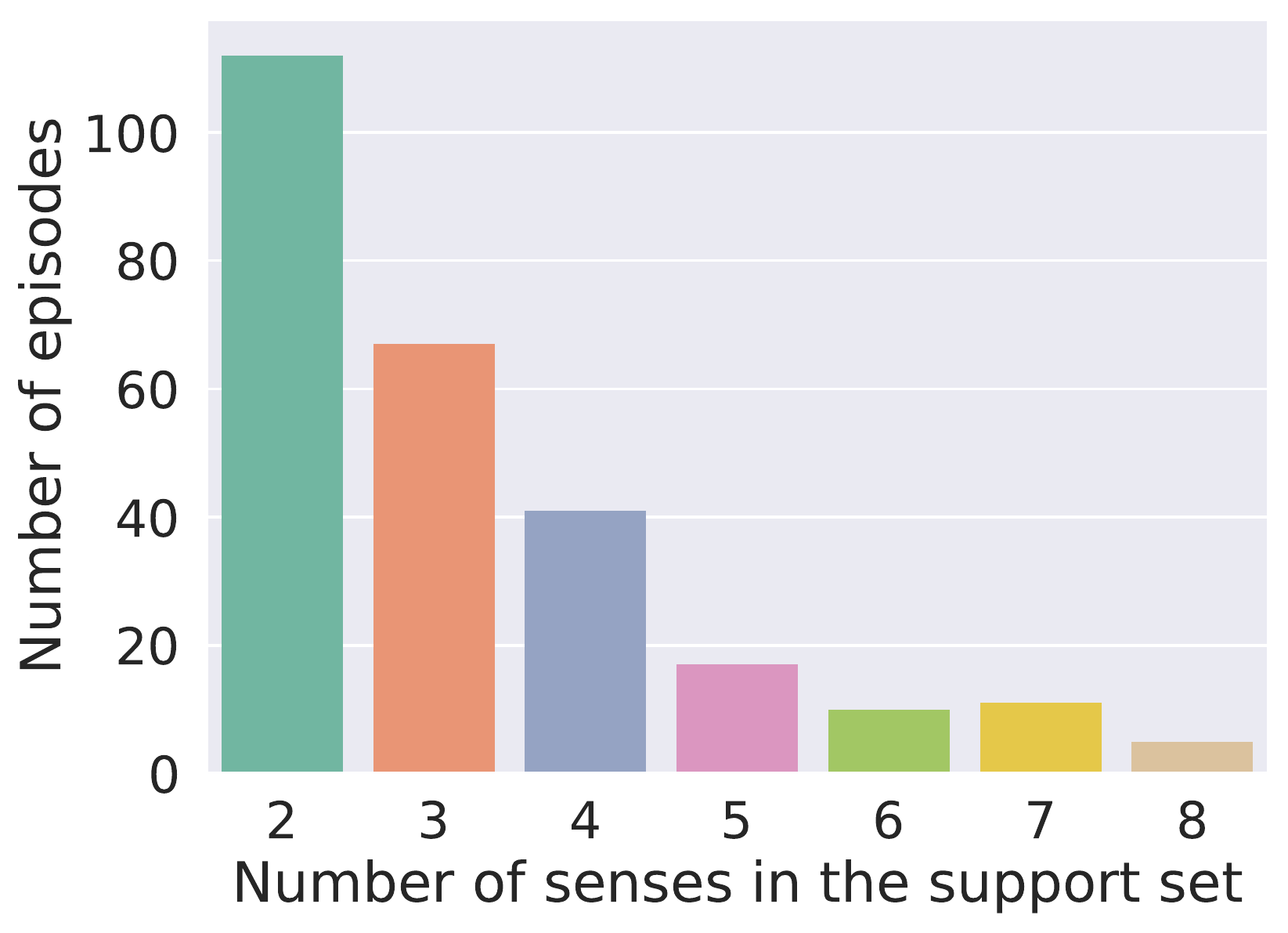}
         \caption{$|S| = 8$}
         \label{fig:sense_sepisode_8}
     \end{subfigure}
     \hfill
     \begin{subfigure}[b]{0.24\textwidth}
         \centering
         \includegraphics[width=\textwidth]{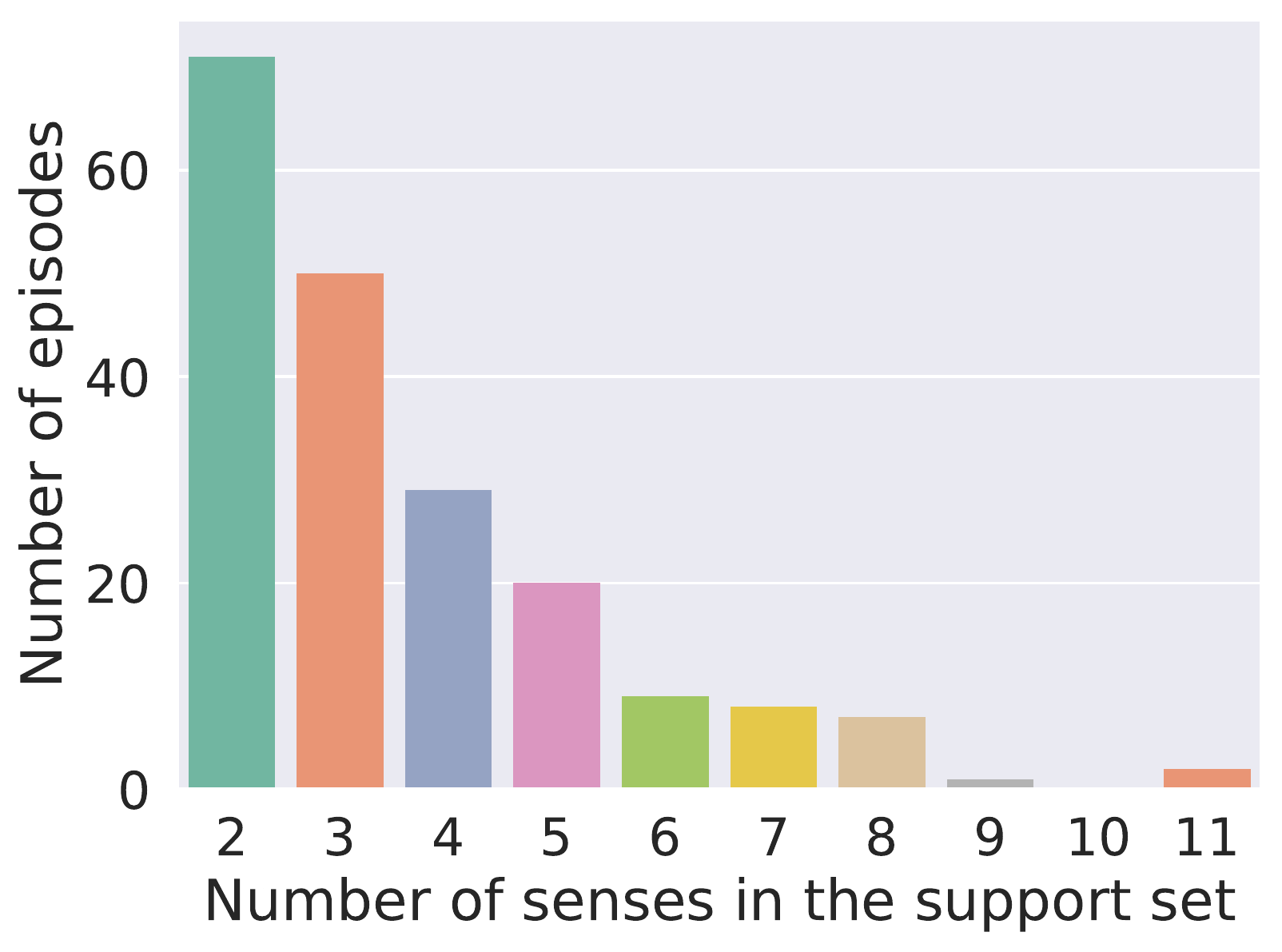}
         \caption{$|S| = 16$}
         \label{fig:sense_sepisode_16}
     \end{subfigure}
     \hfill
     \begin{subfigure}[b]{0.24\textwidth}
         \centering
         \includegraphics[width=\textwidth]{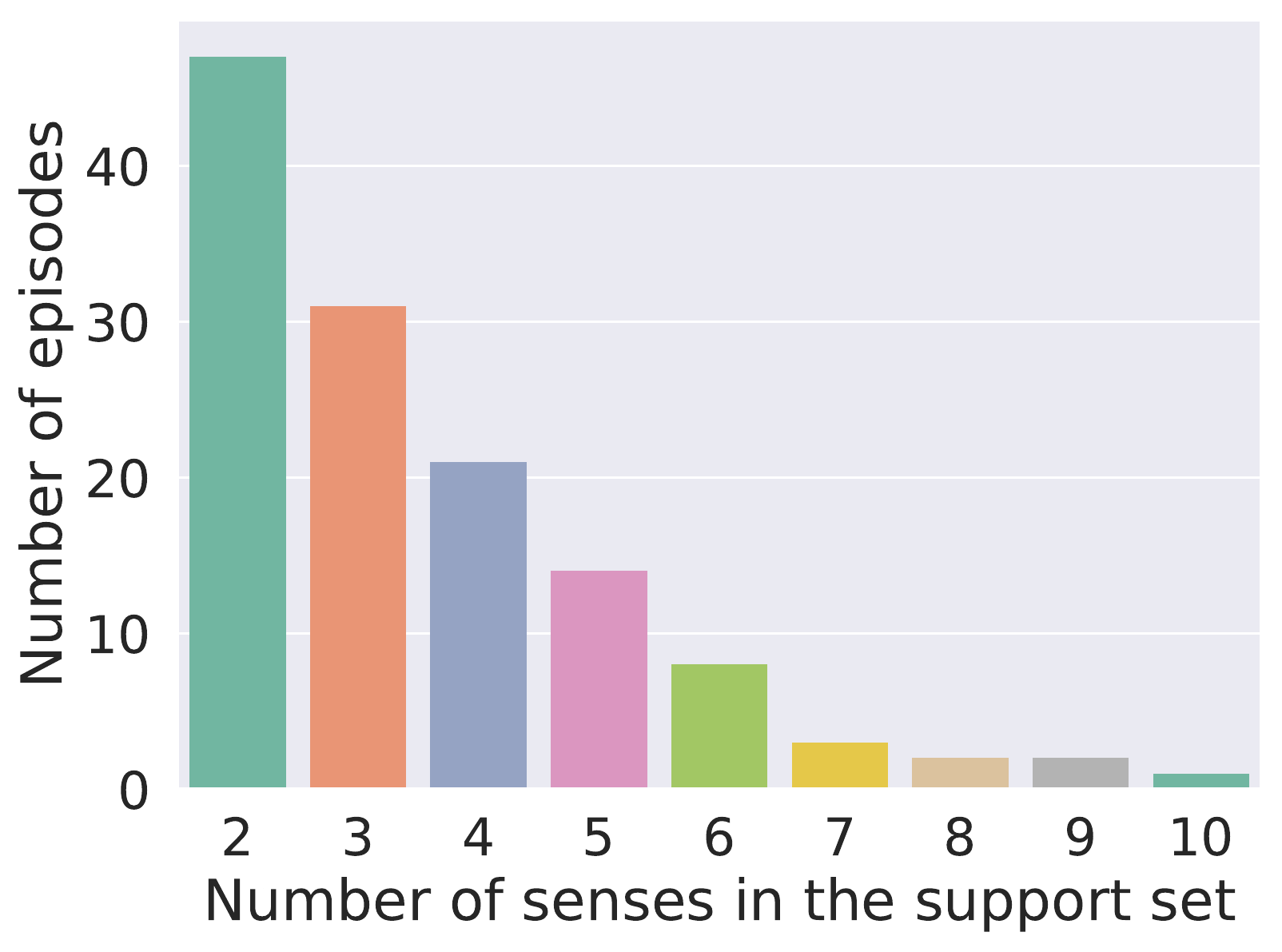}
         \caption{$|S| = 32$}
         \label{fig:sense_sepisode_32}
     \end{subfigure}
    \caption{Bar plot of number of meta-test episodes for different number of senses in the meta-test support set.}
    \label{fig:sense_sepisode_dist}
\end{figure*}

\begin{figure*}[ht]
    \centering
    \begin{subfigure}[b]{0.24\textwidth}
         \centering
         \includegraphics[width=\textwidth]{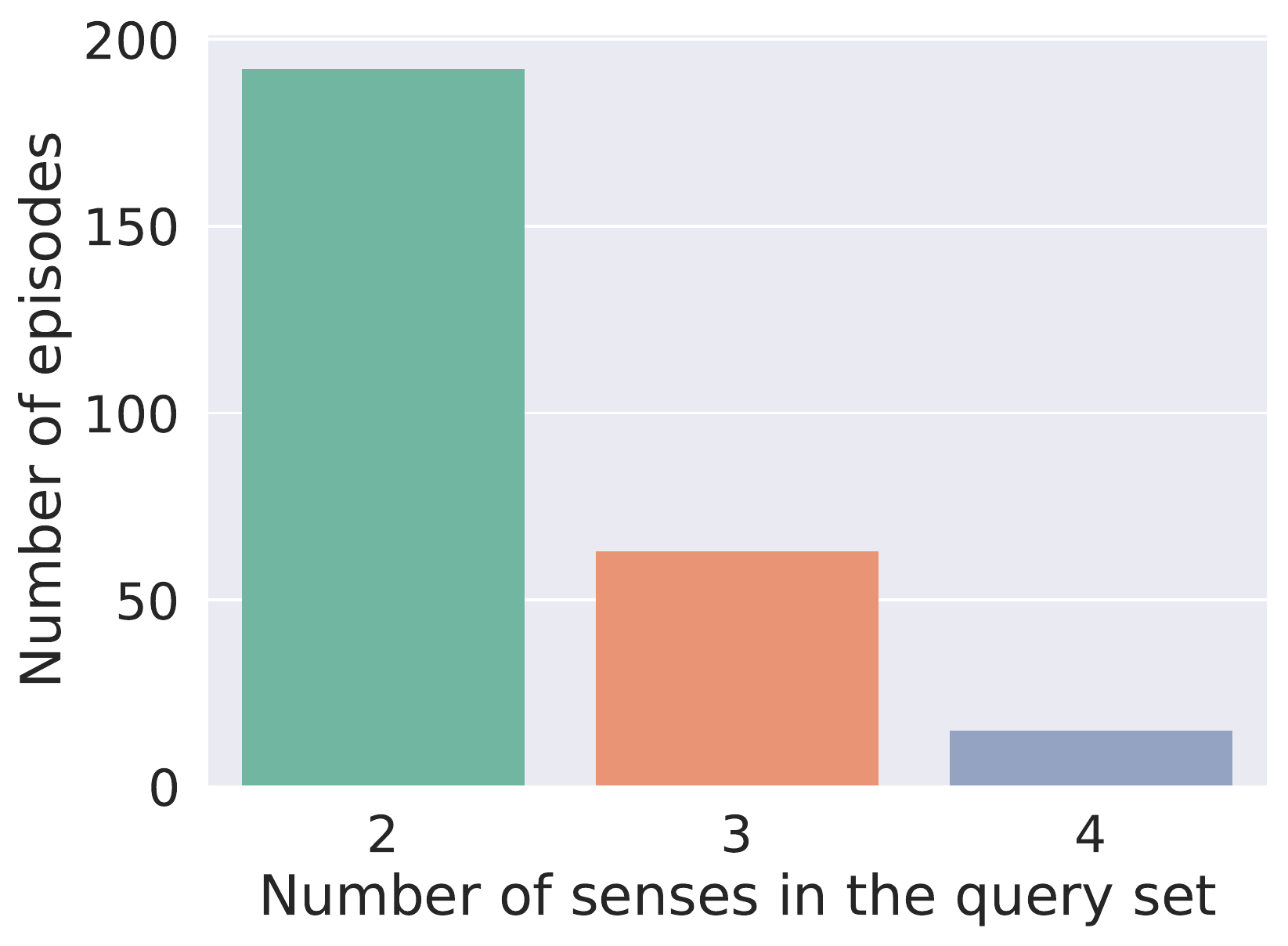}
         \caption{$|S| = 4$}
         \label{fig:sense_qepisode_4}
     \end{subfigure}
     \hfill
    \begin{subfigure}[b]{0.24\textwidth}
         \centering
         \includegraphics[width=\textwidth]{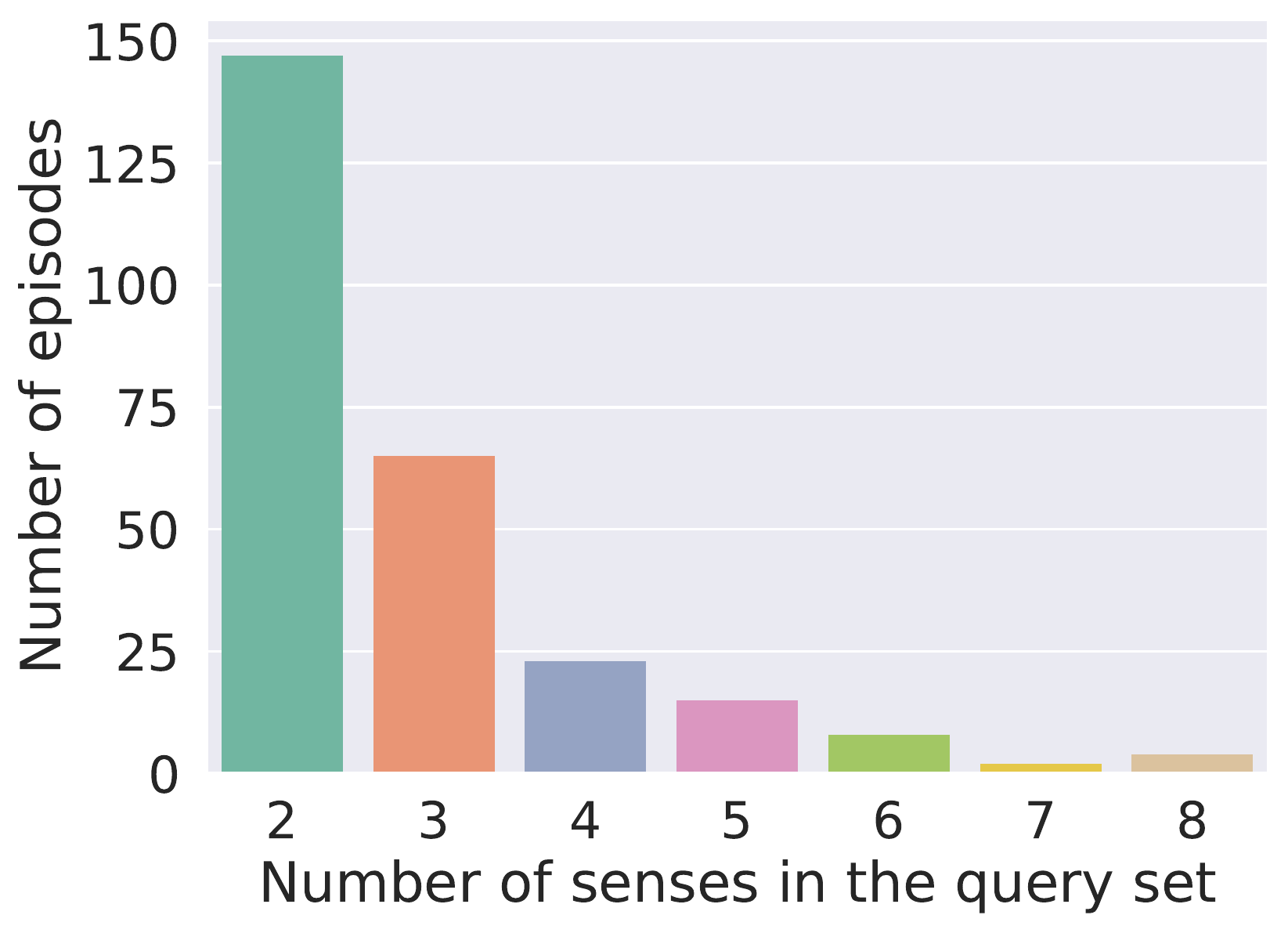}
         \caption{$|S| = 8$}
         \label{fig:sense_qepisode_8}
     \end{subfigure}
     \hfill
     \begin{subfigure}[b]{0.24\textwidth}
         \centering
         \includegraphics[width=\textwidth]{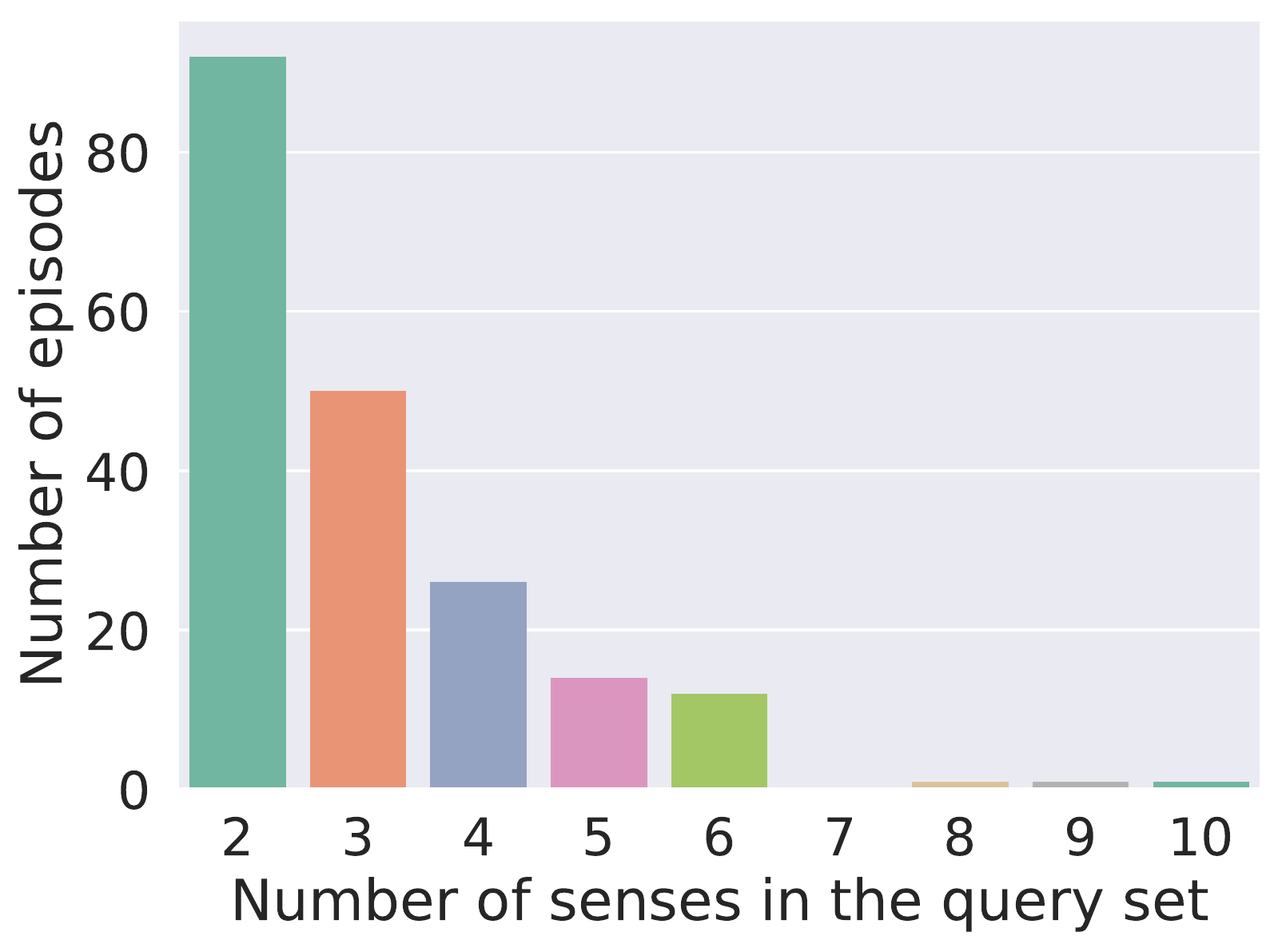}
         \caption{$|S| = 16$}
         \label{fig:sense_qepisode_16}
     \end{subfigure}
     \hfill
     \begin{subfigure}[b]{0.24\textwidth}
         \centering
         \includegraphics[width=\textwidth]{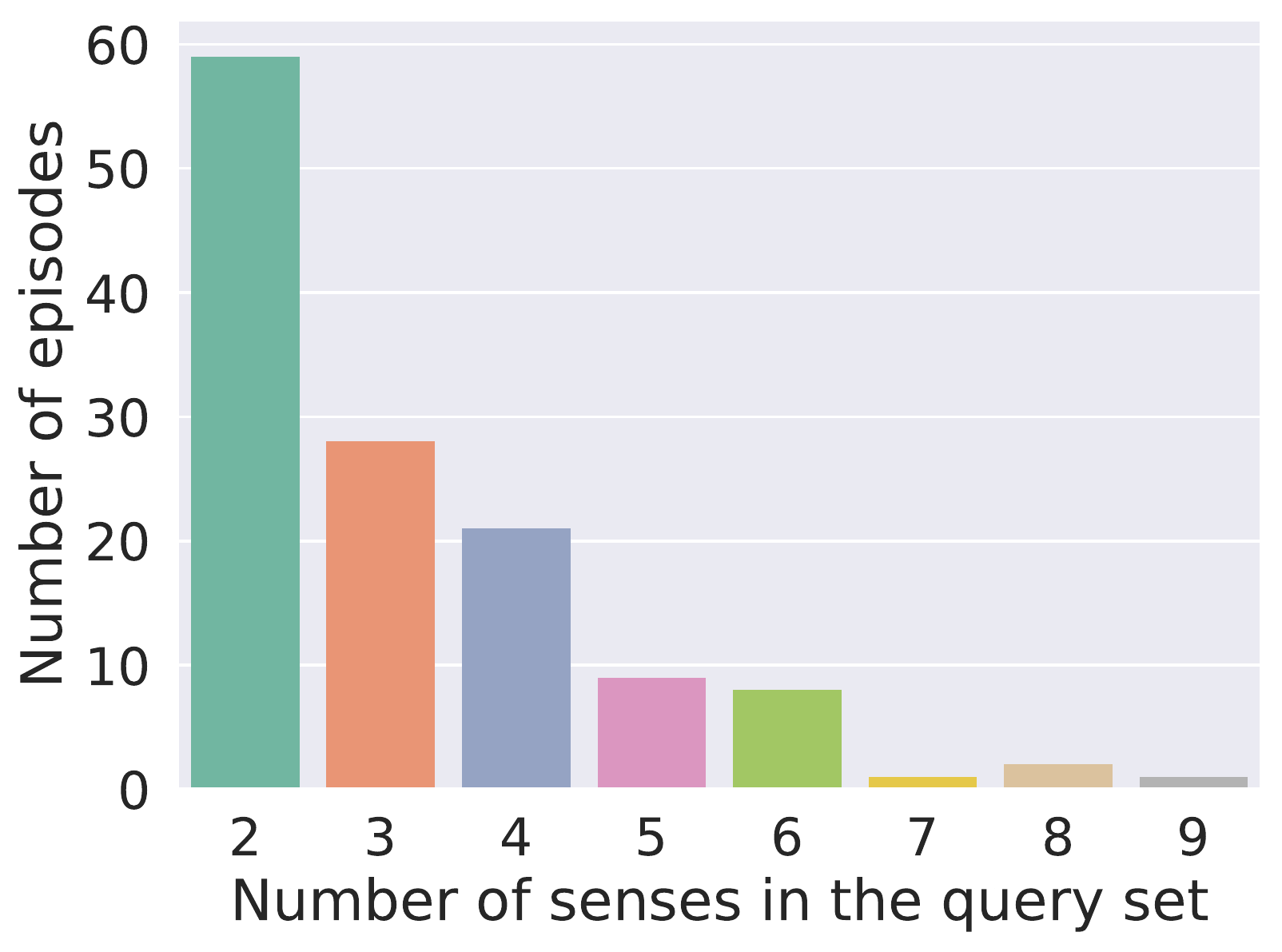}
         \caption{$|S| = 32$}
         \label{fig:sense_qepisode_32}
     \end{subfigure}
    \caption{Bar plot of number of meta-test episodes for different number of senses in the meta-test query set.}
    \label{fig:sense_qepisode_dist}
\end{figure*}

We report the number of words, the number of episodes, the total number of unique sentences and the average number of senses for the meta-training, meta-validation and meta-test sets for each of the four setups with different $|S|$ in Table \ref{tab:data_split}. Additionally, in Figure \ref{fig:sense_sepisode_dist} and Figure \ref{fig:sense_qepisode_dist}, we present bar plots of the number of meta-test episodes for different number of senses in the meta-test support and query sets respectively. It shows that the number of episodes drops quite sharply as the number of senses increases. In each episode, only words with a maximum of $|S|$ senses are considered so that all of them are accommodated in the support set. 

\subsection{Hyperparameters} \label{sec:hyp}

\begin{table*}[ht]
\small
\centering
\begin{tabular}{@{}cccccccc@{}}
\toprule
\makecell{\textbf{Embedding/} \\\textbf{Encoder}} & \makecell{\textbf{Method}} & \makecell{\textbf{Output} \\\textbf{learning} \\\textbf{rate}} & \makecell{\textbf{Learner} \\\textbf{learning} \\ \textbf{rate}} & \makecell{\textbf{Meta} \\\textbf{learning} \\\textbf{rate}} & \makecell{\textbf{Hidden} \\\textbf{size}} & \makecell{\textbf{No. of} \\\textbf{inner-loop} \\\textbf{updates}} & \makecell{\textbf{Size of} \\\textbf{shared} \\\textbf{linear layer}} \\ \midrule
\multirow{5}{*}{GloVe+GRU} &  NE-Baseline & $\expnumber{1}{-1}$ & $\expnumber{5}{-4}$ & -- & 256 & 5 & 64 \\
 &  ProtoNet & -- & -- & $\expnumber{1}{-3}$ & 256 & -- & 64 \\
 &  FOMAML & $\expnumber{1}{-1}$ & $\expnumber{1}{-2}$ & $\expnumber{1}{-3}$ & 256 & 5 & 64 \\
 &  ProtoFOMAML & $\expnumber{1}{-3}$ & $\expnumber{1}{-3}$ & $\expnumber{1}{-3}$ & 256 & 5 & 64 \\
 &  ProtoMAML & $\expnumber{1}{-3}$ & $\expnumber{1}{-3}$ & $\expnumber{1}{-3}$ & 256 & 5 & 64 \\ \midrule
 \multirow{5}{*}{ELMo+MLP} &  NE-Baseline & $\expnumber{1}{-1}$ & $\expnumber{1}{-3}$ & -- & -- & 7 & 256 \\
 &  ProtoNet & -- & -- & $\expnumber{1}{-3}$ & -- & -- & 256 \\
 &  FOMAML & $\expnumber{1}{-1}$ & $\expnumber{1}{-2}$ & $\expnumber{5}{-3}$ & -- & 7 & 256 \\
 &  ProtoFOMAML & $\expnumber{1}{-3}$ & $\expnumber{1}{-3}$ & $\expnumber{5}{-4}$ & -- & 7 & 256 \\
 &  ProtoMAML & $\expnumber{1}{-3}$ & $\expnumber{1}{-3}$ & $\expnumber{5}{-4}$ & -- & 7 & 256 \\ \midrule
 \multirow{4}{*}{BERT} &  NE-Baseline & $\expnumber{1}{-1}$ & $\expnumber{5}{-5}$ & -- & -- & 7 & 192 \\
 &  ProtoNet & -- & -- & $\expnumber{1}{-6}$ & -- & -- & 192 \\
 &  FOMAML & $\expnumber{1}{-1}$ & $\expnumber{1}{-3}$ & $\expnumber{5}{-5}$ & -- & 7 & 192 \\
 &  ProtoFOMAML & $\expnumber{1}{-3}$ & $\expnumber{1}{-3}$ & $\expnumber{1}{-4}$ & -- & 7 & 192 \\ 
\bottomrule
\end{tabular}
\caption{Hyperparameters used for training the models.}
\label{tab:hyperparams}
\end{table*}

We performed hyperparameter tuning for all the models under the $|S|=16$ setting. We obtain the best hyperparameters on the basis of the average macro F1 score on the meta-validation set. We trained the models with five seeds (42 - 46) and recorded the mean of the metric from the five runs to identify the best hyperparameters. For $|S| = 4, 8, 32$, we chose the best hyperparameters obtained from this tuning. 

We employed early stopping with a patience of $2$ epochs, i.e., we stop meta-training if the validation metric does not improve over $2$ epochs. Tuning over all the hyperparameters of our models is prohibitively expensive. Hence, for some of the hyperparameters we chose a fixed value. The size of the shared linear layer is $64$, $256$ and $192$ for the GloVe+GRU, ELMo+MLP and BERT models respectively. The shared linear layer's activation function is \textit{tanh} for GloVe+GRU and \textit{ReLU} for ELMo+MLP and BERT. For FOMAML, ProtoFOMAML and ProtoMAML, the batch size is set to $16$ tasks. For the BERT models, we perform learning rate warm-up for $100$ steps followed by a constant rate. For GloVe+GRU and ELMo+MLP, we decay the learning rate by half every $500$ steps. We also experimented with two types of regularization -- dropout for the inner-loop updates and weight decay for the outer-loop updates -- but both of them yielded a drop in performance. 

The remaining hyperparameters, namely the output learning rate, learner learning rate, meta learning rate, hidden size (only for GloVe+GRU), and number of inner-loop updates were tuned. The search space for these is as follows:
\begin{itemize}
    \itemsep0em 
    \item Output learning rate: $\expnumber{1}{-1}$, $\expnumber{1}{-2}$, $\expnumber{5}{-3}$, $\expnumber{1}{-3}$
    \item Learner learning rate: $\expnumber{1}{-1}$,  $\expnumber{5}{-2}$, $\expnumber{1}{-2}$, $\expnumber{5}{-3}$, $\expnumber{1}{-3}$, $\expnumber{5}{-4}$, $\expnumber{1}{-4}$, $\expnumber{5}{-5}$, $\expnumber{1}{-5}$
    \item Meta learning rate: $\expnumber{5}{-3}$, $\expnumber{1}{-3}$, $\expnumber{5}{-4}$, $\expnumber{1}{-4}$, $\expnumber{5}{-5}$, $\expnumber{1}{-5}$, $\expnumber{5}{-6}$, $\expnumber{1}{-6}$
    \item Hidden size: $64$, $128$, $256$
    \item Number of inner-loop updates: $3$, $5$, $7$
\end{itemize}
The best hyperparameters obtained are shown in Table \ref{tab:hyperparams}.

\subsection{Training times} \label{sec:train_time}

\begin{table*}[ht]
\small
\centering
\begin{tabular}{@{}cccc@{}}
\toprule
\makecell{\textbf{Embedding/} \\\textbf{Encoder}} & \makecell{\textbf{Method}} & \makecell{\textbf{No. of GPUs} \\ \textbf{used}} & \makecell{\textbf{Approximate} \\ \textbf{training time} \\ \textbf{per epoch}} \\ \midrule
\multirow{5}{*}{GloVe+GRU} &  NE-Baseline & 1 & 8 minutes \\
&  ProtoNet & 1 & 8 minutes \\
&  FOMAML & 1 & 15 minutes \\
&  ProtoFOMAML & 1 & 18 minutes \\
&  ProtoMAML & 1 & 9 hours 30 minutes \\ \midrule
\multirow{5}{*}{ELMo+MLP} & NE-Baseline & 1 & 55 minutes \\
&  ProtoNet & 1 & 55 minutes \\
&  FOMAML & 1 & 1 hour \\
&  ProtoFOMAML & 1 & 1 hour \\
&  ProtoMAML & 1 & 1 hour 2 minutes \\ \midrule
\multirow{6}{*}{BERT} & NE-Baseline & 1 & 35 minutes \\
&  ProtoNet & 1 & 35 minutes \\
&  FOMAML & 4 & 2 hours 35 minutes \\
&  ProtoFOMAML & 4 & 4 hours 18 minutes \\
&  ProtoFOMAML* & 1 & 41 minutes \\
&  ProtoMAML* & 1 & 47 minutes \\
\bottomrule
\end{tabular}
\caption{Approximate training time per epoch. (*Only the top layer fine-tuned and only for one inner-loop step.)}
\label{tab:train_time}
\end{table*}

We train all our models on TitanRTX GPUs. Our model architectures vary in the total number of trainable parameters. Thus, the time taken to train each of them varies. The number of meta-learned parameters $\bm{\theta}$ is as follows:
\begin{itemize}
    \itemsep0em 
    \item GloVe+GRU: $889,920$
    \item ELMo+MLP: $262,404$
    \item BERT: $107,867,328$
\end{itemize} 

To give an idea of how long it takes to train them, we provide the approximate time taken for one epoch for the $|S|=16$ setup in Table \ref{tab:train_time}. The training time would be slightly lower for $|S| = 4, 8$ and slightly higher for $|S|=32$. The training time for ProtoMAML with GloVe+GRU is extremely long (second-order derivatives for RNNs with the cuDNN backend is not supported in PyTorch and hence cuDNN had to be disabled). 

\subsection{Challenging cases} \label{sec:challenging_cases}

In Table \ref{tab:worst_words}, we present $10$ words with the lowest macro F1 scores (in increasing order of the score) obtained from ProtoNet with GloVe+GRU. We perform the analysis on this model to investigate challenging cases without the contextualization advantage offered by ELMo and BERT. For $|S| = 4, 8, 16$, many words in the list have predominantly verb senses, showing that they are more challenging to disambiguate. The number of such cases drops in $|S| = 32$, indicating that disambiguation of verbs improves as $|S|$ increases.

\subsection{F1 score distribution} \label{sec:f1_dist}

\begin{figure*}[ht]
    \centering
    \begin{subfigure}[b]{0.24\textwidth}
         \centering
         \includegraphics[width=\textwidth]{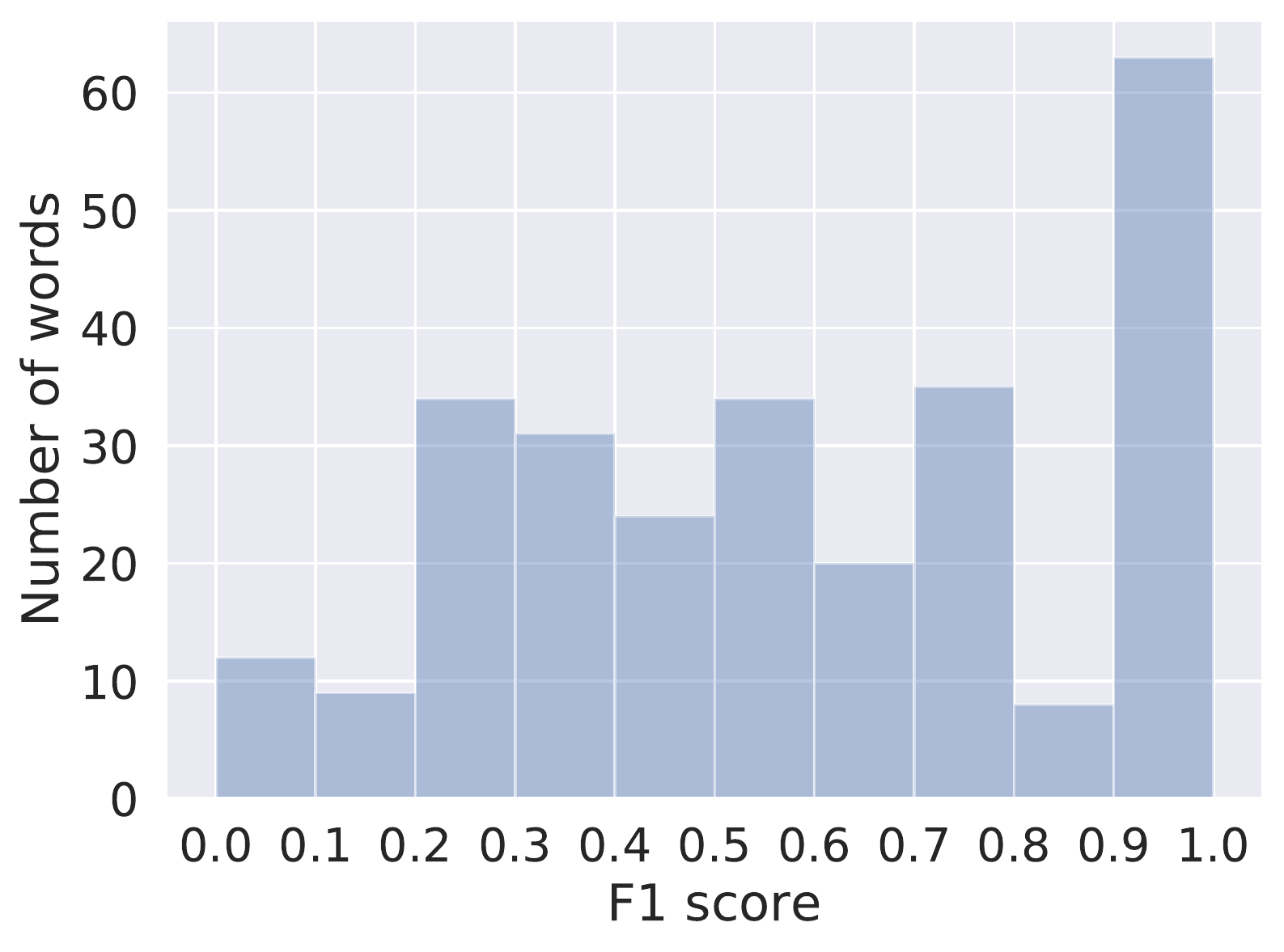}
         \caption{$|S| = 4$}
         \label{fig:f1_hist_4}
     \end{subfigure}
     \hfill
    \begin{subfigure}[b]{0.24\textwidth}
         \centering
         \includegraphics[width=\textwidth]{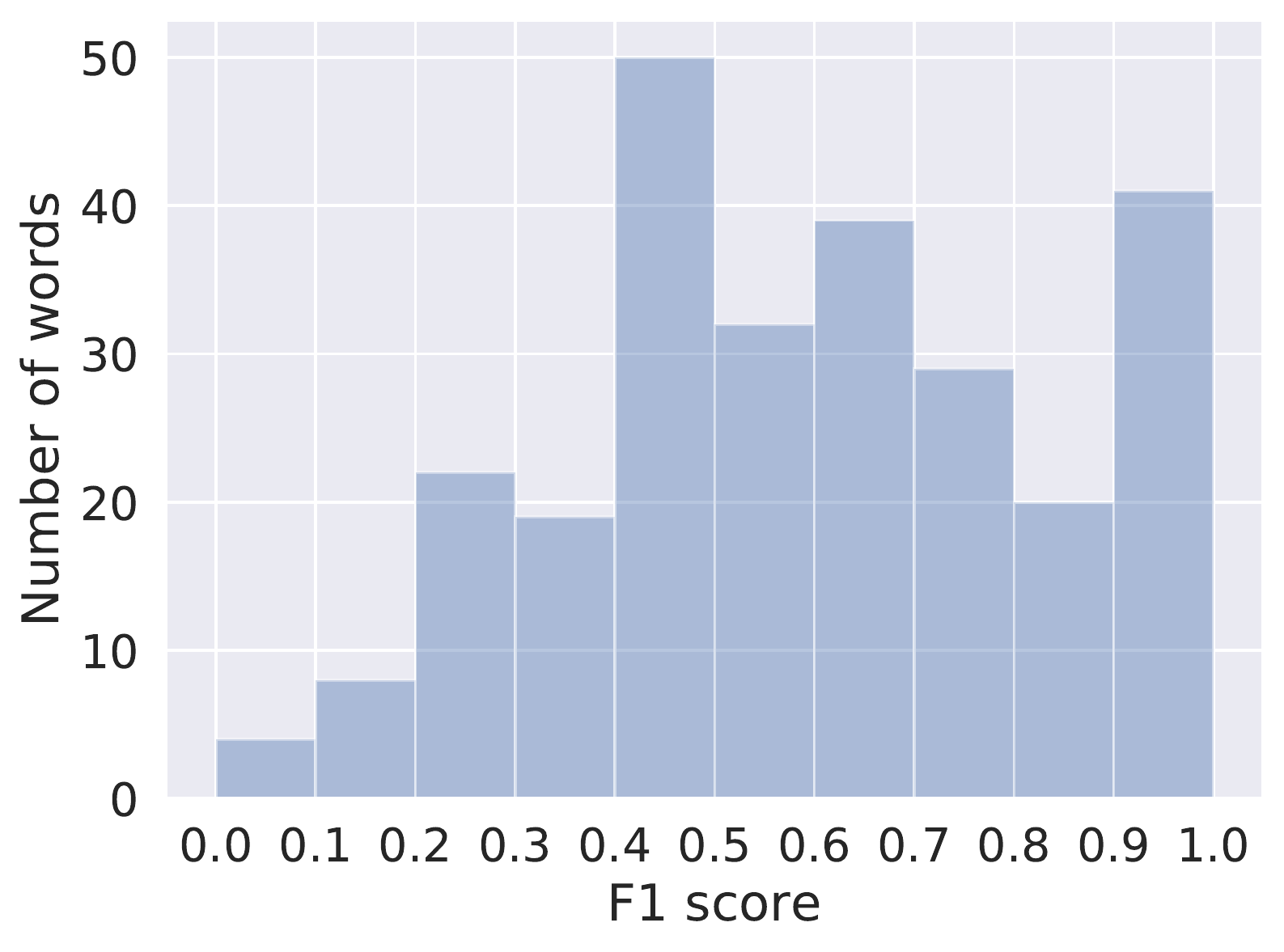}
         \caption{$|S| = 8$}
         \label{fig:f1_hist_8}
     \end{subfigure}
     \hfill
     \begin{subfigure}[b]{0.24\textwidth}
         \centering
         \includegraphics[width=\textwidth]{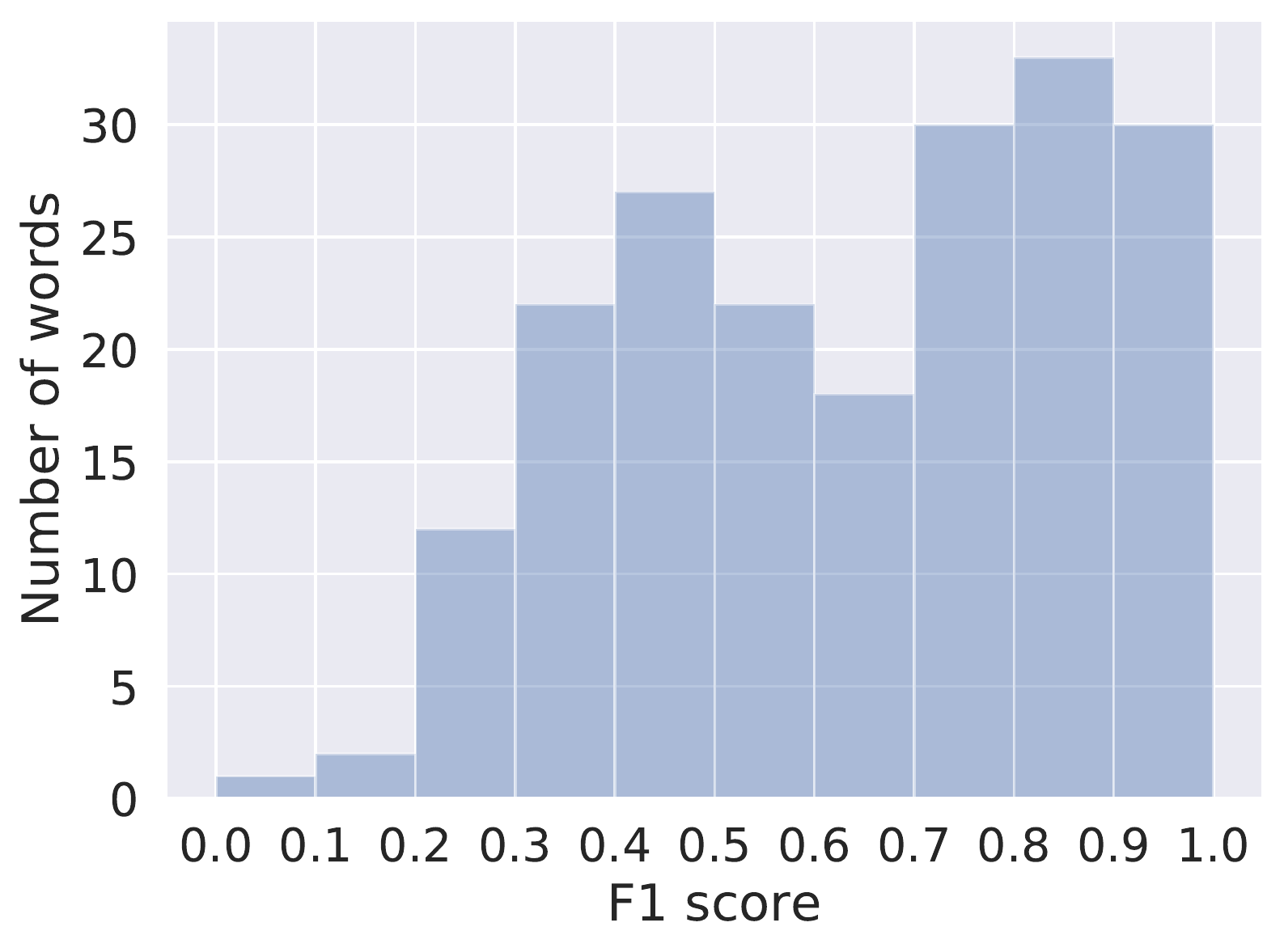}
         \caption{$|S| = 16$}
         \label{fig:f1_hist_16}
     \end{subfigure}
     \hfill
     \begin{subfigure}[b]{0.24\textwidth}
         \centering
         \includegraphics[width=\textwidth]{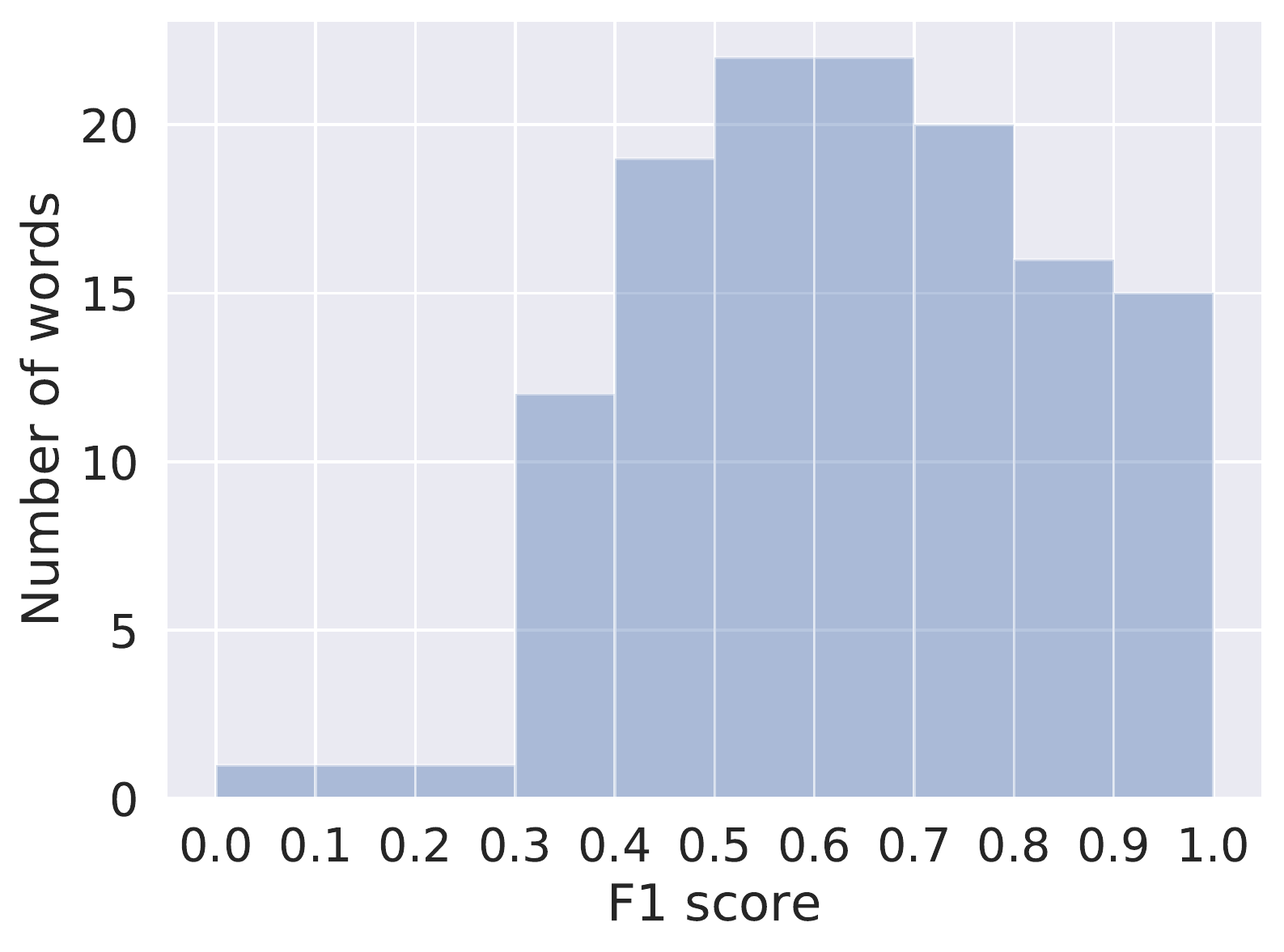}
         \caption{$|S| = 32$}
         \label{fig:f1_hist_32}
     \end{subfigure}
    \caption{Distribution of macro F1 scores for ProtoNet with GloVe+GRU.}
    \label{fig:f1_hist}
\end{figure*}

For ProtoNet with GloVe+GRU, we plot the distribution of macro F1 scores across the words in the meta-test set in Figure \ref{fig:f1_hist}. The distribution is mostly right-skewed with very few words having scores in the range $0$ to $0.2$. 

\subsection{t-SNE visualizations} \label{sec:tsne}

\begin{figure*}[ht]
    \centering
    \begin{subfigure}[b]{0.32\textwidth}
         \centering
         \includegraphics[width=\textwidth]{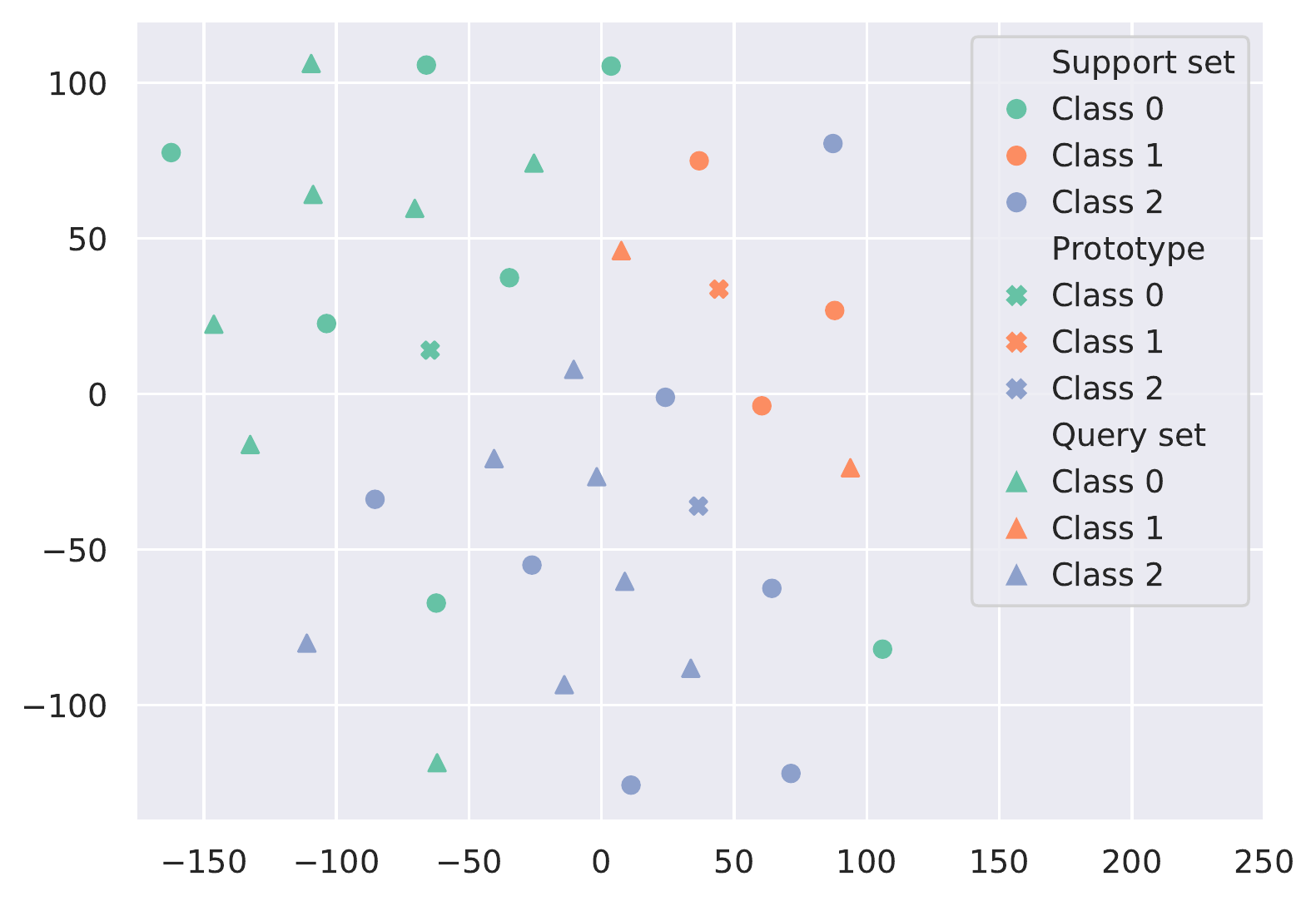}
         \caption{Word representation -- `favor'}
         \label{fig:proto_glove_visualize_favor}
     \end{subfigure}
     \hfill
     \begin{subfigure}[b]{0.32\textwidth}
         \centering
         \includegraphics[width=\textwidth]{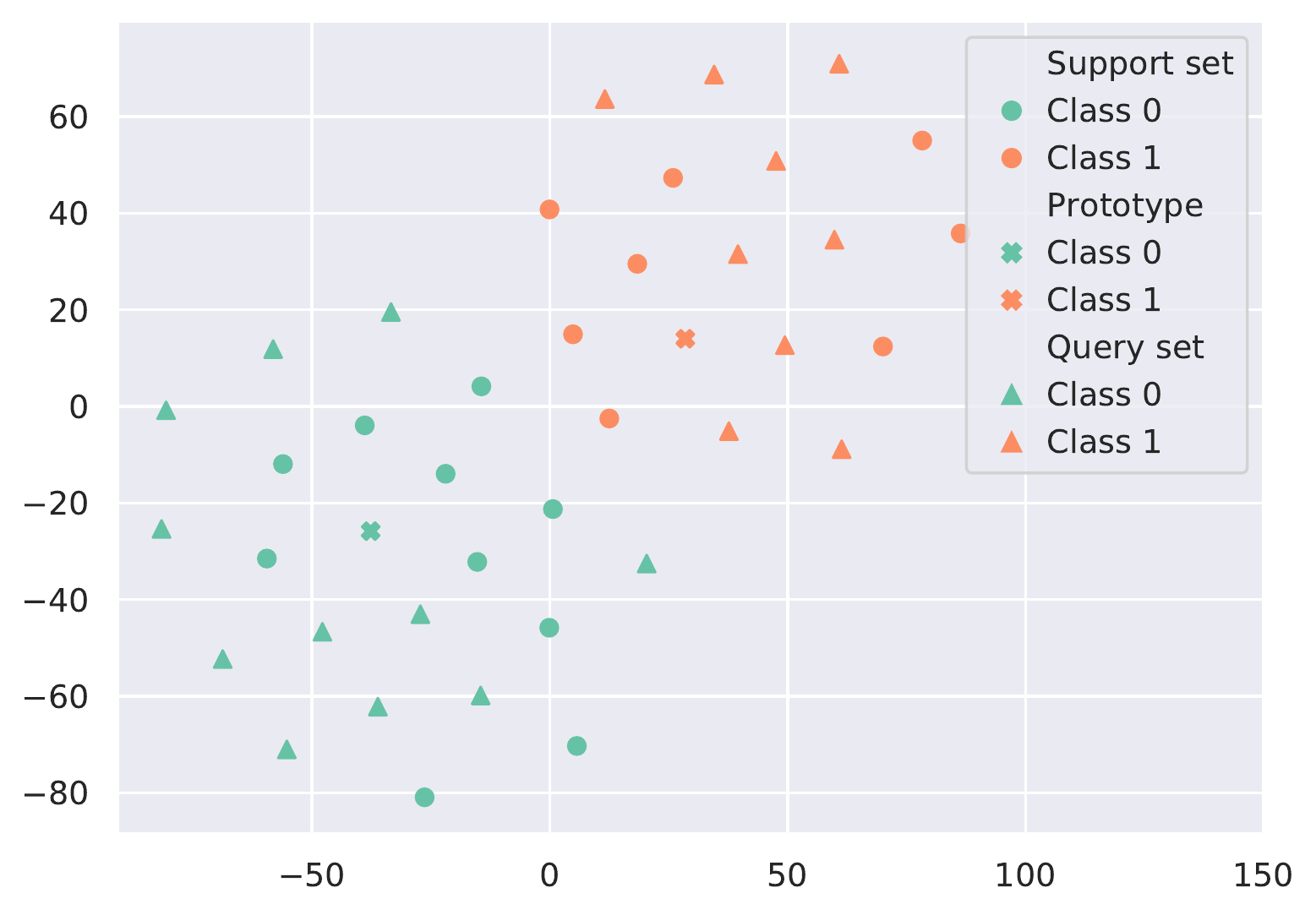}
         \caption{Word representation -- `tax'}
         \label{fig:proto_glove_visualize_tax}
     \end{subfigure}
     \hfill
     \begin{subfigure}[b]{0.32\textwidth}
         \centering
         \includegraphics[width=\textwidth]{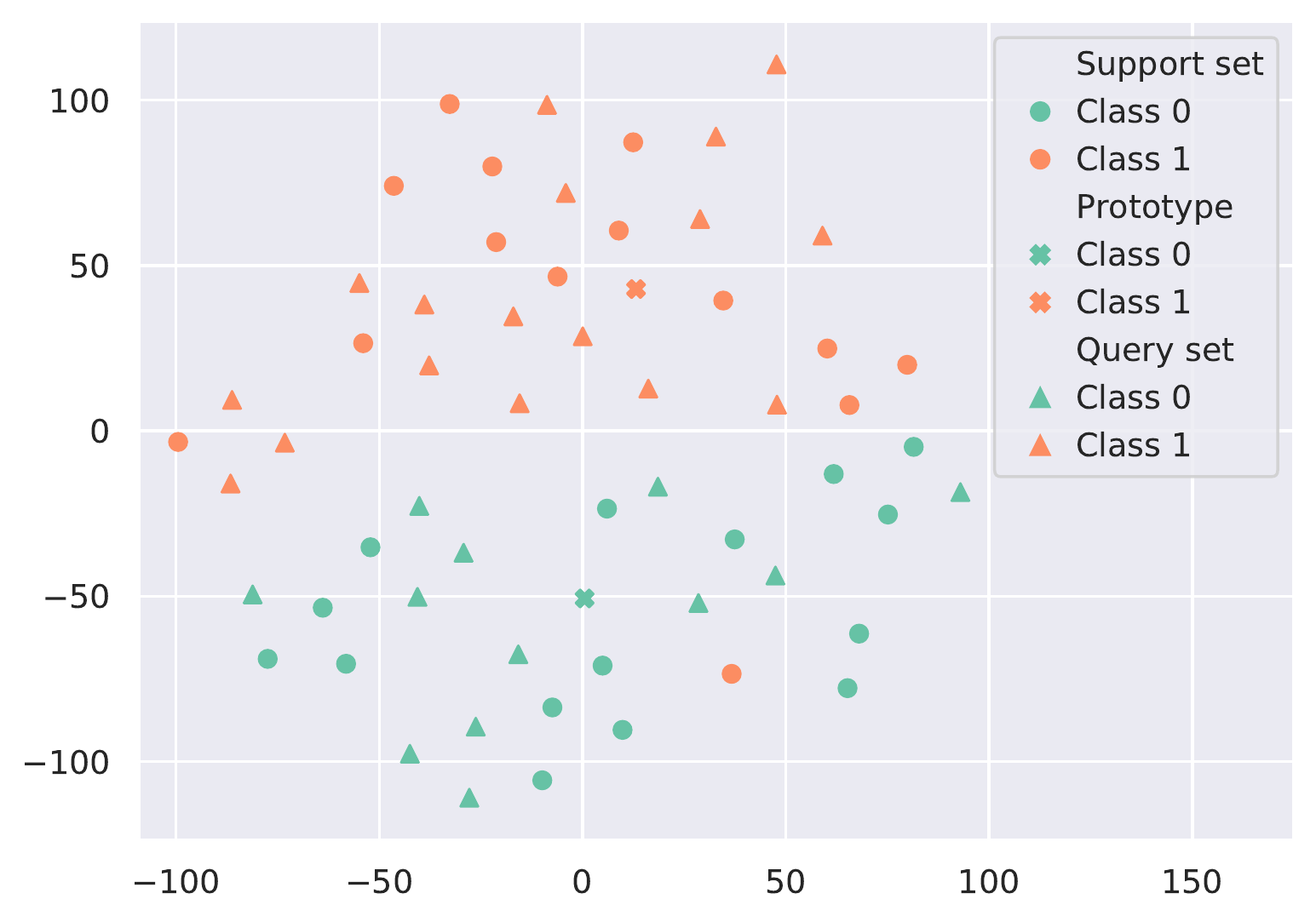}
         \caption{Word representation -- `snow'}
         \label{fig:proto_glove_visualize_snow}
     \end{subfigure}
    \caption{t-SNE visualizations of word representations generated by ProtoNet with GloVe+GRU.}
    \label{fig:tsne_glove}
\end{figure*}

We provide t-SNE visualizations of the word representations generated by $f_{\bm{\theta}}$ of ProtoNet with GloVe+GRU for three words (with macro F1 score of $1$) in the meta-test set in Figure \ref{fig:tsne_glove}. Even though it receives the same input embedding for all senses, it manages to separate the senses into clusters on the basis of the representations of the support examples. This occurs even though ProtoNet does not perform any fine-tuning step on the support set. Moreover, the query examples also seem to be part of the same cluster and lie close to the prototypes.

\begin{figure*}[ht]
    \centering
    \begin{subfigure}[b]{0.32\textwidth}
         \centering
         \includegraphics[width=\textwidth]{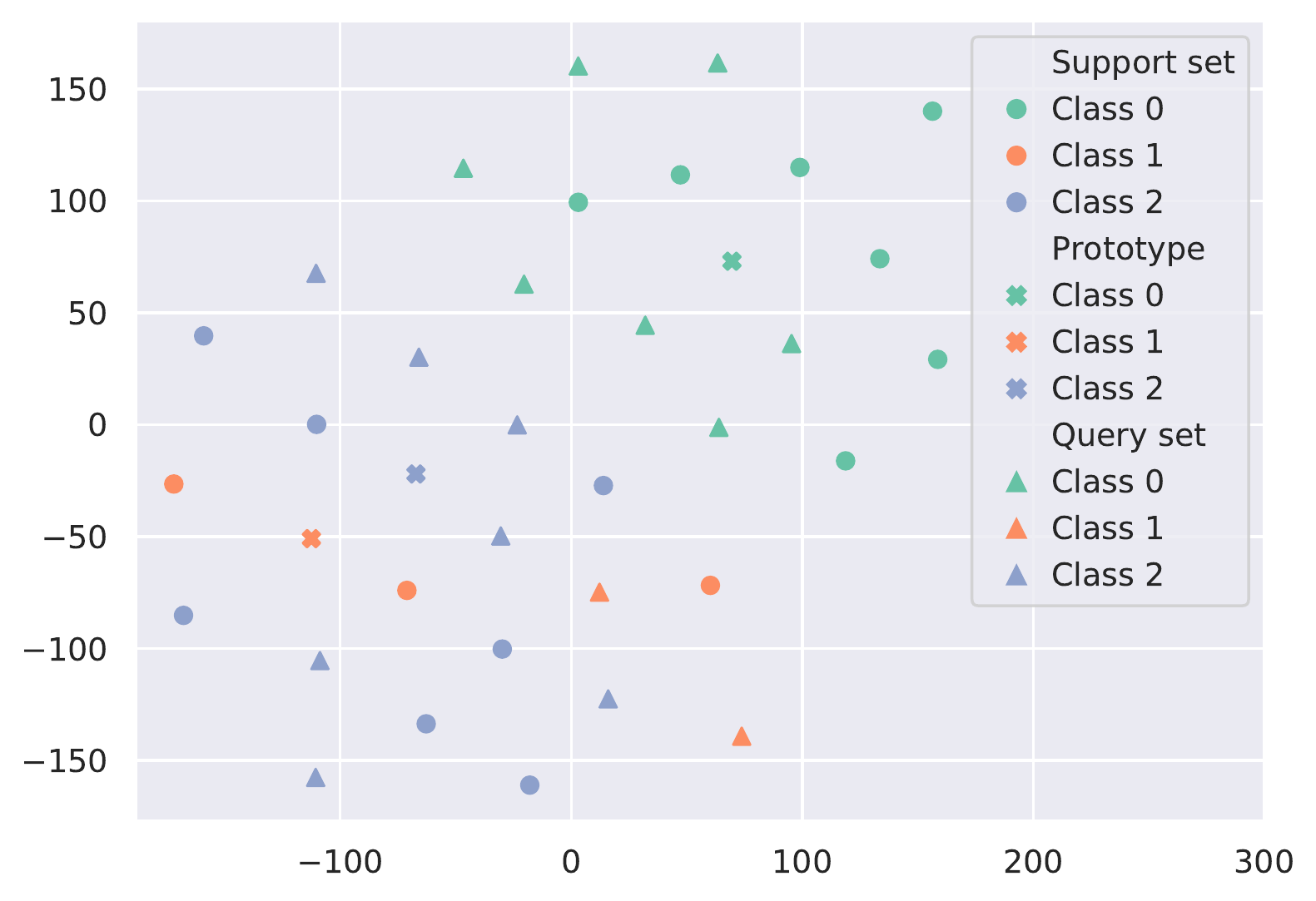}
         \caption{ELMo embedding -- `favor'}
         \label{fig:elmo_visualize_favor}
     \end{subfigure}
     \hfill
     \begin{subfigure}[b]{0.32\textwidth}
         \centering
         \includegraphics[width=\textwidth]{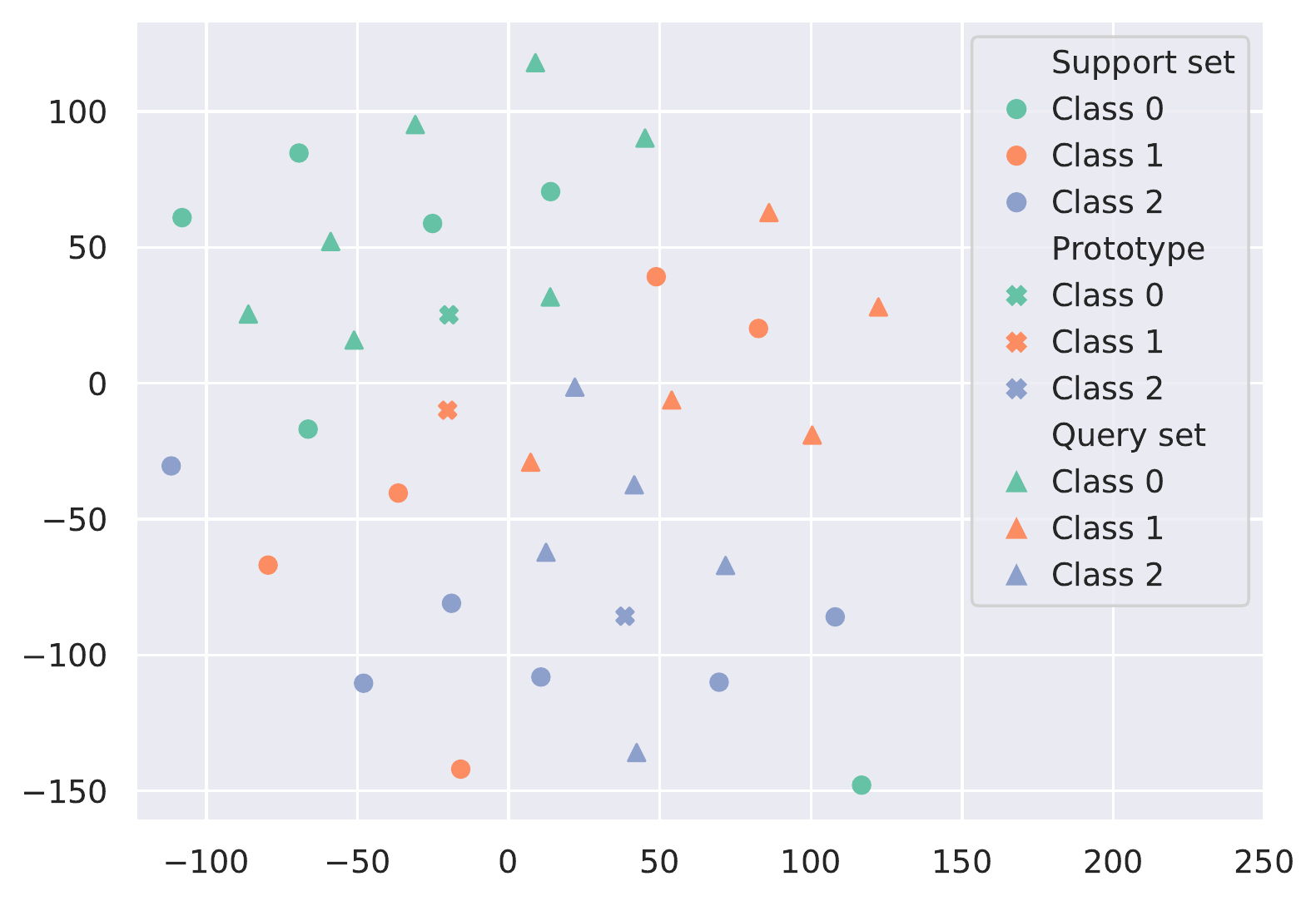}
         \caption{ELMo embedding -- `field'}
         \label{fig:elmo_visualize_field}
     \end{subfigure}
     \hfill
     \begin{subfigure}[b]{0.32\textwidth}
         \centering
         \includegraphics[width=\textwidth]{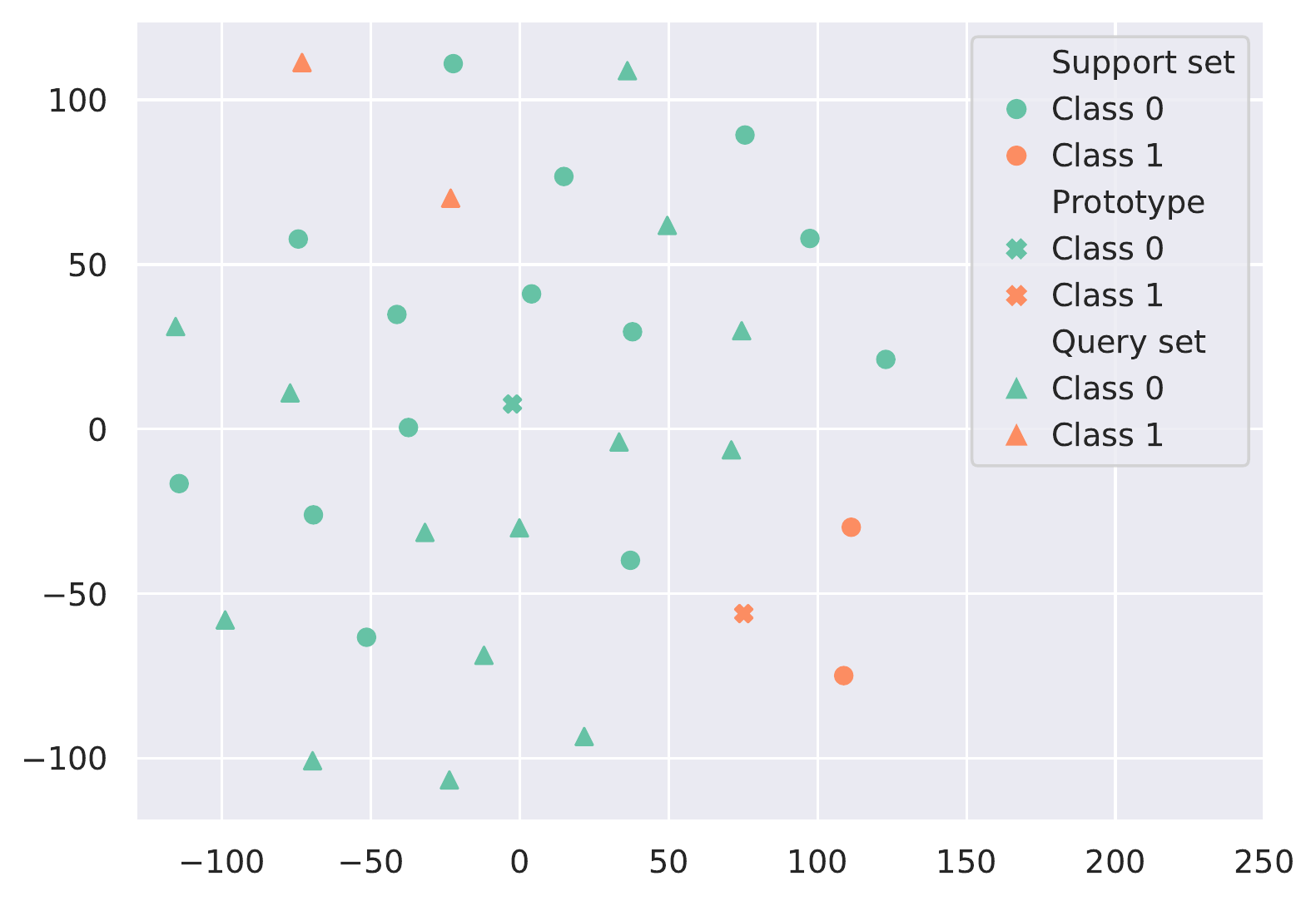}
         \caption{ELMo embedding -- `factor'}
         \label{fig:elmo_visualize_factor}
     \end{subfigure} \\
    \begin{subfigure}[b]{0.32\textwidth}
         \centering
         \includegraphics[width=\textwidth]{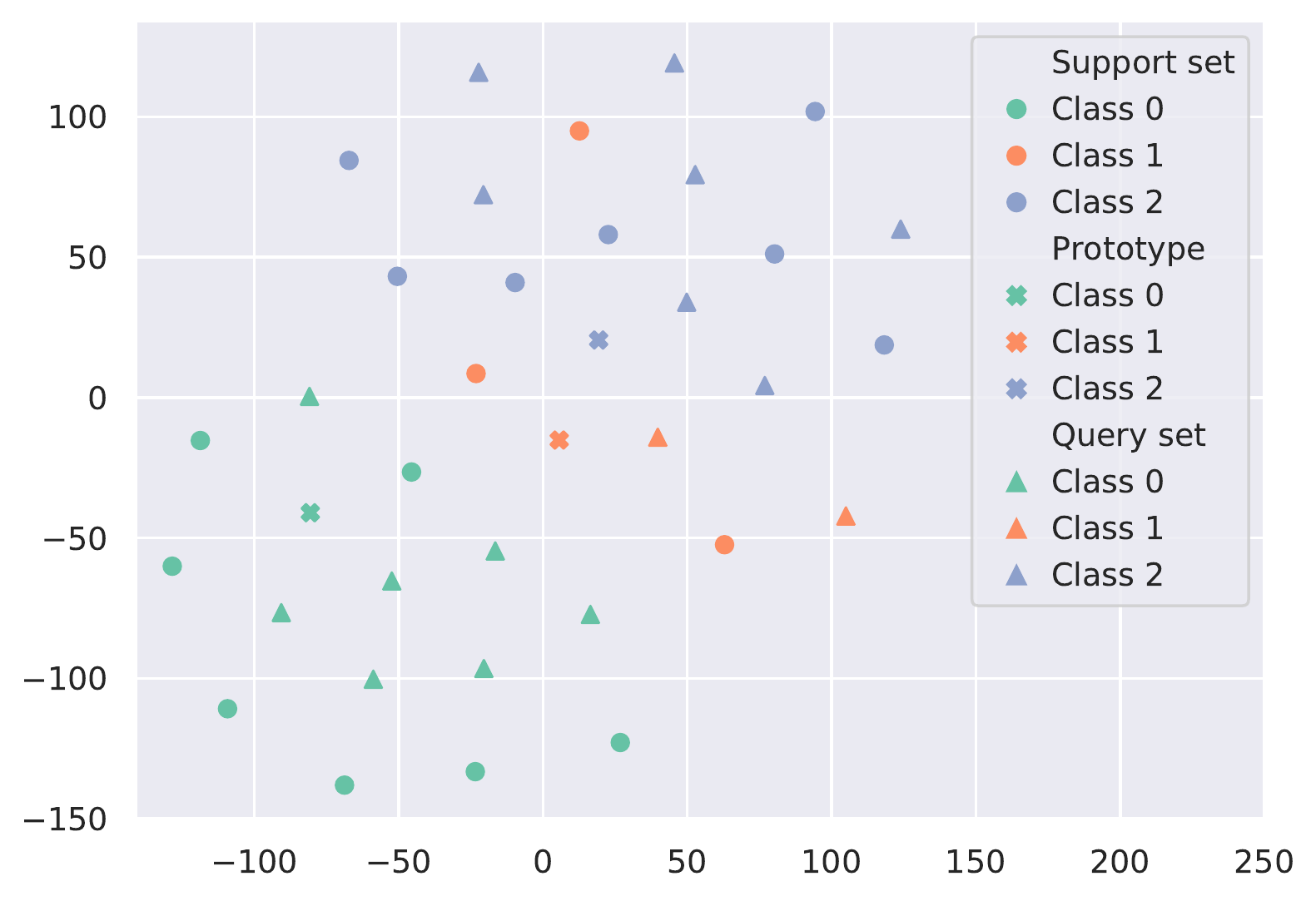}
         \caption{Word representation -- `favor'}
         \label{fig:proto_elmo_visualize_favor}
     \end{subfigure}
     \hfill
     \begin{subfigure}[b]{0.32\textwidth}
         \centering
         \includegraphics[width=\textwidth]{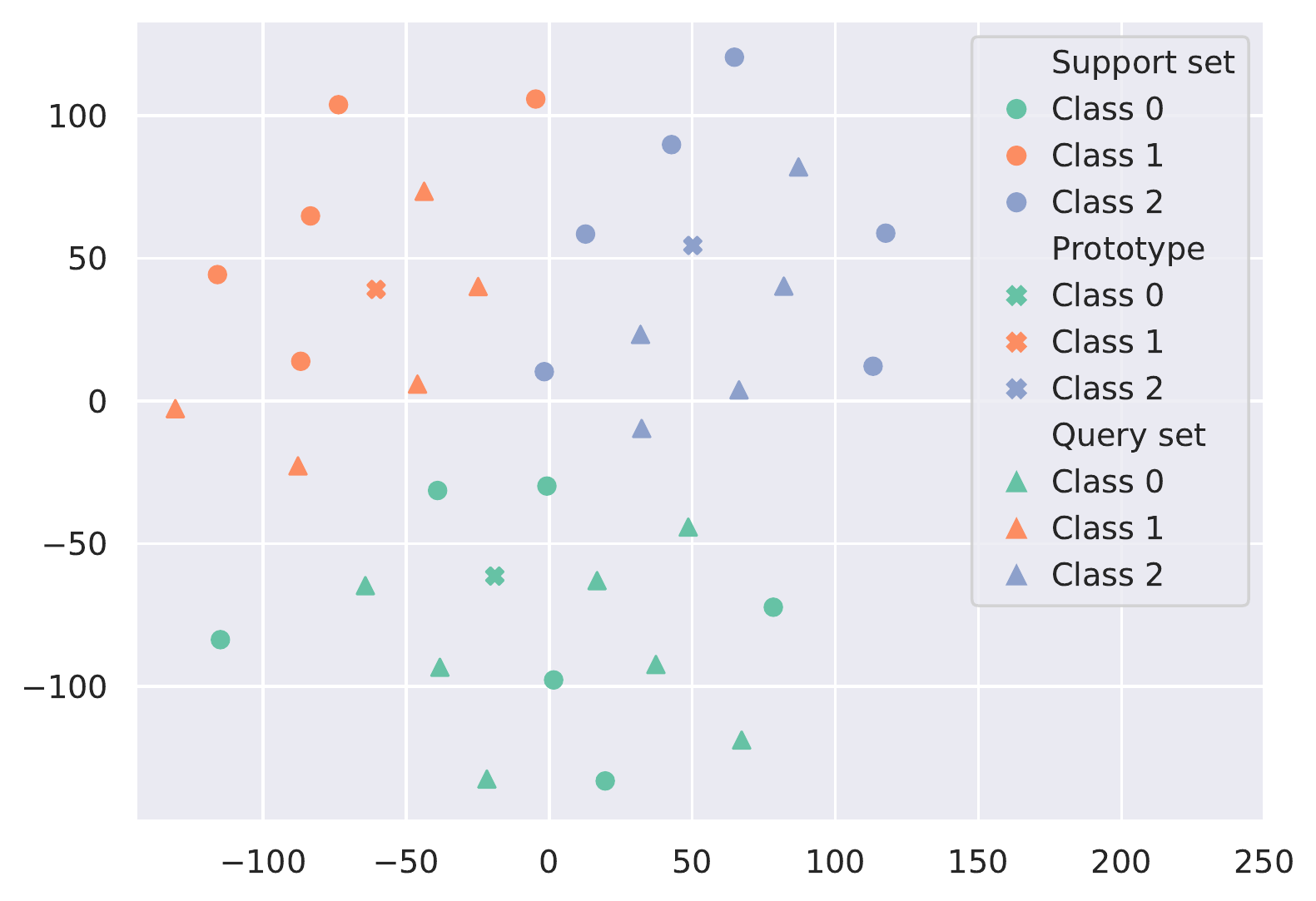}
         \caption{Word representation -- `field'}
         \label{fig:proto_elmo_visualize_field}
     \end{subfigure}
     \hfill
     \begin{subfigure}[b]{0.32\textwidth}
         \centering
         \includegraphics[width=\textwidth]{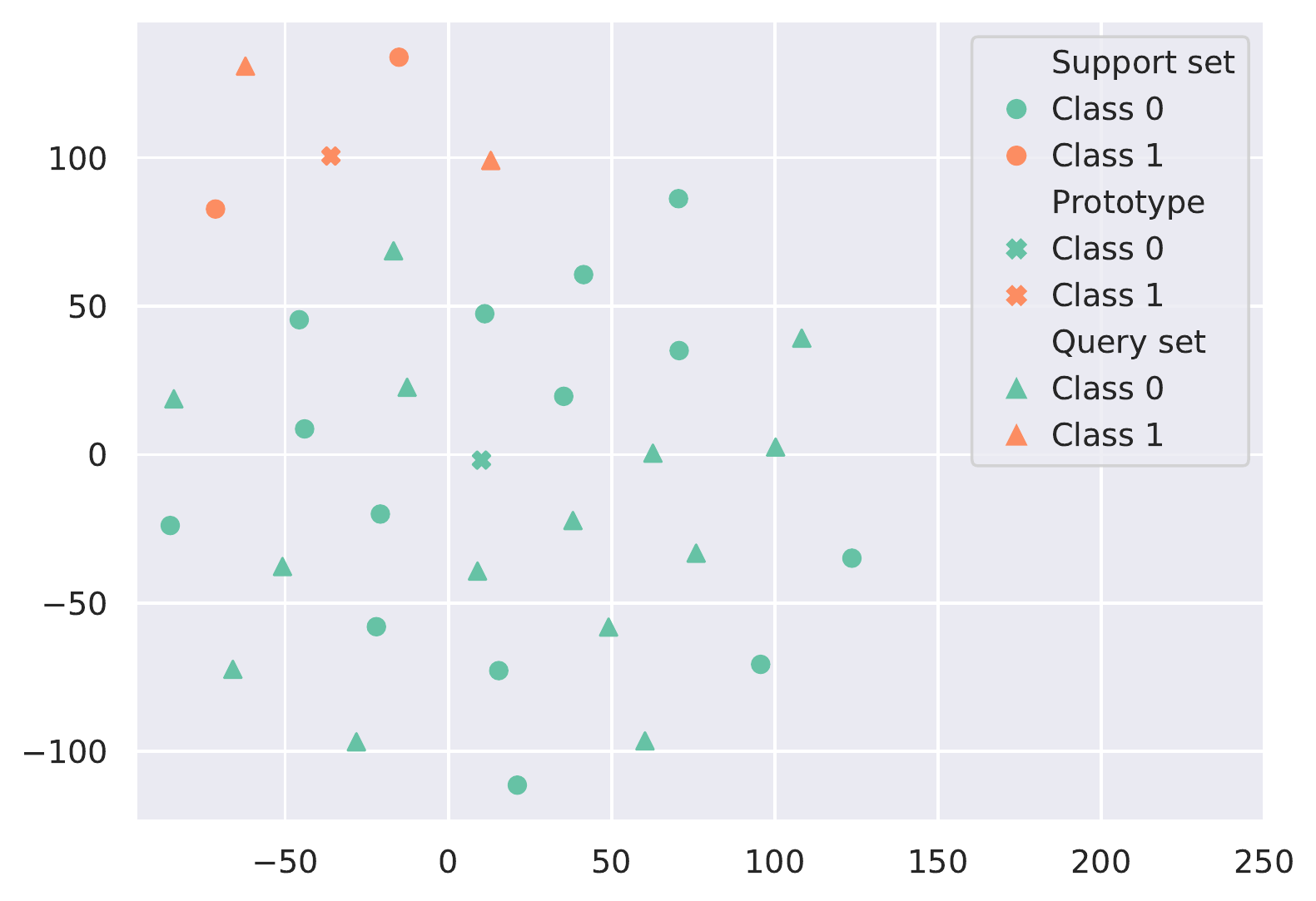}
         \caption{Word representation -- `factor'}
         \label{fig:proto_elmo_visualize_factor}
     \end{subfigure} 
    \caption{t-SNE visualizations comparing ELMo embeddings (top) against word representations generated by ProtoNet with ELMo+MLP (bottom).}
    \label{fig:tsne_elmo}
\end{figure*}

ELMo embeddings, being contextual, already capture information in how the various senses are represented. In order to compare them against the representations generated by ProtoNet with ELMo+MLP, we again provide t-SNE visualizations. We plot the ELMo embeddings of three words in the meta-test test in Figure \ref{fig:elmo_visualize_favor}, \ref{fig:elmo_visualize_field} and \ref{fig:elmo_visualize_factor}. We also show the prototypes computed from these embeddings for illustration. For the same three words, we plot the representations obtained from $f_{\bm{\theta}}$ of ProtoNet with ELMo+MLP in Figure \ref{fig:proto_elmo_visualize_favor}, \ref{fig:proto_elmo_visualize_field} and \ref{fig:proto_elmo_visualize_factor}. It can be observed that the ELMo embeddings alone are not well-clustered with respect to the senses. On the other hand, ProtoNet manages to separate the senses into clusters, which aids in making accurate predictions on the query set. 

These visualizations further demonstrate ProtoNet's success in disambiguating new words. From a learning to learn standpoint, the model has learned how to separate the senses in a high-dimensional space so as to disambiguate them. Proto(FO)MAML often improves upon this good initialization during its inner-loop updates.

\subsection{Results on the meta-validation set}

To facilitate reproducibility, we provide the results on the meta-validation set for all the methods that involved hyperparameter tuning in Table \ref{tab:results_val}. 

\begin{table*}[ht]
\small
\centering
\begin{tabular}{llllll}
\toprule
\multirow{2}{*}{\makecell{\textbf{Embedding/} \\ \textbf{Encoder}}} & \multicolumn{1}{c}{\multirow{2}{*}{\textbf{Method}}} &  \multicolumn{4}{c}{\textbf{Average macro F1 score}} \\
                       & & \multicolumn{1}{c}{$|S| = 4$} & \multicolumn{1}{c}{$|S| = 8$}  & \multicolumn{1}{c}{$|S| = 16$} & \multicolumn{1}{c}{$|S| = 32$} \\ \midrule
\multirow{5}{*}{GloVe+GRU} & NE-Baseline & 0.557 $\pm$ 0.015 & 0.563 $\pm$ 0.011 & 0.590 $\pm$ 0.008 & 0.541 $\pm$ 0.018 \\
& ProtoNet & 0.591 $\pm$ 0.008 & 0.615 $\pm$ 0.001 & 0.638 $\pm$ 0.007 & 0.634 $\pm$ 0.006 \\
& FOMAML & 0.540 $\pm$ 0.011 & 0.410 $\pm$ 0.006 & 0.405 $\pm$ 0.007 & 0.351 $\pm$ 0.007 \\
& ProtoFOMAML & \textbf{0.604 $\pm$ 0.016} & \textbf{0.622 $\pm$ 0.010} & \textbf{0.642 $\pm$ 0.005} & 0.626 $\pm$ 0.015 \\
& ProtoMAML & 0.599 $\pm$ 0.004 & \textbf{0.622 $\pm$ 0.010} & 0.641 $\pm$ 0.005 & \textbf{0.627 $\pm$ 0.013} \\\midrule
\multirow{5}{*}{ELMo+MLP} & NE-Baseline &  0.659 $\pm$ 0.016 & 0.685 $\pm$ 0.005 & 0.728 $\pm$ 0.004 & 0.693 $\pm$ 0.007 \\
& ProtoNet & 0.682 $\pm$ 0.008 & 0.701 $\pm$ 0.007 & \textbf{0.741 $\pm$ 0.007} & 0.722 $\pm$ 0.011  \\
& FOMAML & 0.670 $\pm$ 0.007 & 0.609 $\pm$ 0.011 & 0.598 $\pm$ 0.017 & 0.566 $\pm$ 0.011 \\
& ProtoFOMAML & \textbf{0.702 $\pm$ 0.002} & \textbf{0.728 $\pm$ 0.007} & 0.740 $\pm$ 0.003 & 0.732 $\pm$ 0.005 \\
& ProtoMAML & \textbf{0.702 $\pm$ 0.007} & 0.726 $\pm$ 0.008 & \textbf{0.741 $\pm$ 0.003} & \textbf{0.738 $\pm$ 0.006} \\ \midrule
\multirow{6}{*}{BERT} & NE-Baseline & 0.466 $\pm$ 0.160 & 0.601 $\pm$ 0.006 & 0.616 $\pm$ 0.009 & 0.569 $\pm$ 0.006 \\
& ProtoNet & \textbf{0.742 $\pm$ 0.007} & 0.759 $\pm$ 0.013 & 0.786 $\pm$ 0.004 & \textbf{0.770 $\pm$ 0.009} \\
& FOMAML & 0.702 $\pm$ 0.005 & 0.553 $\pm$ 0.019 & 0.506 $\pm$ 0.014 & 0.418 $\pm$ 0.020 \\
& ProtoFOMAML & 0.740 $\pm$ 0.010 & 0.756 $\pm$ 0.008 & \textbf{0.770 $\pm$ 0.009} & 0.734 $\pm$ 0.014  \\
& ProtoFOMAML* & 0.738 $\pm$ 0.016 & \textbf{0.763 $\pm$ 0.003} & 0.769 $\pm$ 0.006 & 0.744 $\pm$ 0.006 \\ 
& ProtoMAML* & 0.737 $\pm$ 0.012 & 0.760 $\pm$ 0.007 & 0.764 $\pm$ 0.005 & 0.736 $\pm$ 0.009 \\ 
\bottomrule                       
\end{tabular}
\caption{Average macro F1 scores of the meta-validation words. (*Only the top layer fine-tuned and for only one inner-loop step)}
\label{tab:results_val}
\end{table*}

\end{document}